%% file: main.tex
\documentclass[twoside]{article}
\usepackage[accepted]{aistats2025}
\usepackage{natbib} % has a nice set of citation styles and commands
\usepackage{booktabs}\usepackage{graphics}
\renewcommand{\bibsection}{\subsubsection*{References}}
\usepackage{multirow}
\usepackage{booktabs}

\usepackage[normalem]{ulem}

\usepackage[utf8]{inputenc} % allow utf-8 input
\usepackage[T1]{fontenc}    % use 8-bit T1 fonts
\usepackage{hyperref}       % hyperlinks
\usepackage{url}            % simple URL typesetting
\usepackage{booktabs}       % professional-quality tables
\usepackage{amsfonts}       % blackboard math symbols
\usepackage{nicefrac}       % compact symbols for 1/2, etc.
\usepackage{microtype}      % microtypography
\usepackage{xcolor}         % colors
\usepackage{makecell}

\input{math_commands.tex}

\usepackage[utf8]{inputenc} % allow utf-8 input
\usepackage[T1]{fontenc}    % use 8-bit T1 fonts
\usepackage{hyperref}       % hyperlinks
\usepackage{url}            % simple URL typesetting
\usepackage{booktabs}       % professional-quality tables
\usepackage{amsfonts}       % blackboard math symbols
\usepackage{nicefrac}       % compact symbols for 1/2, etc.
\usepackage{microtype}      % microtypography
\usepackage{xcolor}         % colors
\usepackage{bbm}

\usepackage{subcaption}
\usepackage{float}
\usepackage{array}
\usepackage{amssymb}
\usepackage{graphicx}
\usepackage{placeins}
\usepackage{amsmath}
\usepackage{mathtools}
\usepackage{amsthm}
\usepackage{thmtools}
\usepackage{thm-restate}
\usepackage{tabularx}
\usepackage{xspace}
\usepackage{cleveref}

\theoremstyle{definition}
\newtheorem{definition}{Definition}[section]

\newtheorem{theorem}{Theorem}[section]
\newtheorem{lemma}{Lemma}[section]
\newtheorem{proposition}{Proposition}[section]

\newcommand{\defeq}{\vcentcolon=}

\newcommand{\lce}{\ell_{\text{CE}}}
\newcommand{\error}{\varepsilon}

\usepackage{xparse}
\NewDocumentCommand{\carl}
{ mO{} }{\textcolor{blue}{\textsuperscript{\textit{Carl}}\textsf{\textbf{\small[#1]}}}}

\NewDocumentCommand{\han}
{ mO{} }{\textcolor{red}{\textsuperscript{\textit{Han}}\textsf{\textbf{\small[#1]}}}}

\NewDocumentCommand{\akul}
{ mO{} }{\textcolor{magenta}{\textsuperscript{\textit{Akul}}\textsf{\textbf{\small[#1]}}}}

\newcommand{\algo}{WaSS}

\newcommand{\Sys}{WaSS \xspace}

\author{
Akul Goyal \\
University of Illinois Urbana-Champaign \\
201 North Goodwin Ave \\
\texttt{akulg2@illinois.edu} \\
\AND
Carl Edwards
University of Illinois Urbana-Champaign \\
201 North Goodwin Ave \\
\texttt{akulg2@illinois.edu} \\
}

\begin{document}
\pagestyle{plain}
\renewcommand{\headrulewidth}{0pt}
\runningauthor{Goyal \& Edwards}
\runningtitle{WaSS}

\twocolumn[

\aistatstitle{Class-based Subset Selection for Transfer Learning\\ under Extreme Label Shift}

\aistatsauthor{ Akul Goyal \\
University of Illinois Urbana-Champaign \\
Department of Computer Science \\
\texttt{akulg2@illinois.edu} \\
\And 
Carl Edwards \\
University of Illinois Urbana-Champaign \\
Department of Computer Science \\
\texttt{cne2@illinois.edu} \\}

\aistatsaddress{}

]

% \maketitle

\input{0abstract}
\input{1introduction}
\input{3methodology}

\input{4results}

\input{2related}
\input{5conclusion}

\FloatBarrier

%https://www.overleaf.com/learn/latex/Bibliography_management_with_natbib
%\bibliographystyle{ksfh_nat}

\clearpage

\bibliographystyle{abbrvnat} %unsrt %plainnat
% \setcitestyle{authoryear,open={((},close={))}}
\bibliography{main}

\clearpage

\appendix

\input{6appendix}

\end{document}

%% file: math_commands.tex
\usepackage{amsmath,amsfonts,bm}

\def\1{\bm{1}}

\def\eps{{\epsilon}}

\DeclareMathAlphabet{\mathsfit}{\encodingdefault}{\sfdefault}{m}{sl}
\SetMathAlphabet{\mathsfit}{bold}{\encodingdefault}{\sfdefault}{bx}{n}

\def\gD{{\mathcal{D}}}

\def\gH{{\mathcal{H}}}

\def\gX{{\mathcal{X}}}
\def\gY{{\mathcal{Y}}}
\def\gZ{{\mathcal{Z}}}

\newcommand{\E}{\mathbb{E}}

\newcommand{\R}{\mathbb{R}}

\DeclareMathOperator*{\argmax}{arg\,max}
\DeclareMathOperator*{\argmin}{arg\,min}

\DeclareMathOperator{\Tr}{Tr}

%% file: 0abstract.tex
\begin{abstract}
Existing work within transfer learning often follows a two-step process -- pre-training over a large-scale source domain and then finetuning over limited samples from the target domain. Yet, despite its popularity, this methodology has been shown to suffer in the presence of distributional shift -- specifically when the output spaces diverge. Previous work has focused on increasing model performance within this setting by identifying and classifying only the shared output classes between distributions. However, these methods are inherently limited as they ignore classes outside the shared class set, disregarding potential information relevant to the model transfer. This paper proposes a new process for few-shot transfer learning that selects and weighs classes from the source domain to optimize the transfer between domains. More concretely, we use Wasserstein distance to choose a set of source classes and their weights that minimize the distance between the source and target domain. To justify our proposed algorithm, we provide a generalization analysis of the performance of the learned classifier over the target domain and show that our method corresponds to a bound minimization algorithm. We empirically demonstrate the effectiveness of our approach (\algo) by experimenting on several different datasets and presenting superior performance within various label shift settings, including the extreme case where the label spaces are disjoint. 

\end{abstract}

%% file: 1introduction.tex
\section{Introduction}
As machine learning becomes increasingly data-dependent, it is more efficient to leverage existing labeled datasets when adapting to novel scenarios. However, large publicly available datasets are often not distributionally representative of the specialized task. For example, images in Imagenet ~\citep{deng2009imagenet} or COCO ~\citep{lin2014microsoft} are largely different from images captured by a niche sensor - negatively impacting a model trained on both datasets ~\citep{pang2018zoom}. Transfer learning, particularly in domain adaptation, has become a popular approach to reducing this distributional shift and enabling knowledge to be transferred more easily between datasets. Previous work ~\citep{zhang2020domain, zhang2019neural} in few-shot learning has shown that it is possible to learn invariant features over the source and target datasets using a few labeled samples from the target distribution~\citep{li2021learning}, facilitating a more accurate transfer. 

 Other work has shown success in reducing the distributional shift within the input space~\citep{Ganin,zhao2018adversarial}; however, these methods struggle when the distributional shift occurs within the label space~\citep{zhao2019domain,johansson2019support,zhao2022fundamental}. In many cases, prior methods' performances degrade due to negative transfer~\citep{wang2019characterizing,zhao2019domain}, defined by the increase in the \emph{risk} in the target distribution between a model trained on the combination of source and target distributions versus it solely trained on the target distribution. In other words, given similarity exists in the output space, divergent classes (classes that only appear in the source distribution) reduce the transferability of a model to the target distribution.

Motivated by the above observation, we aim to reduce negative transfer in this work by eliciting the optimal subset of source classes that allows for the best transfer to the target domain. Contrary to previous methods, our work makes no assumptions about the relationship between the class sets of the source and target domain. Our approach can, therefore, handle extreme cases even when no overlapping classes exist between the two domains - which we refer to this setting as Disjoint Set Domain Adaptation (DDA) for the rest of the paper. Instead of manually training and testing against every possible subset of source class combinations, we propose to use a proxy to recover the optimal subset of source classes efficiently. Previous work~\citep{courty2017joint} has shown that Wasserstein distance can be used to measure divergence between domains. Consequently, we adapt Wasserstein distance to class selection, creating a sparse reweighting that is the most amenable to transfer.

We implement our algorithm using a generalized backbone model (Resnet-50 pre-trained on ImageNet) to map data from a source/target domain into the same feature space where similarity is easily measured ~\citep{norouzi2013zero}. We then apply Wasserstein distance within the embedding space, recovering the class weighting. To measure transferability, we train a neural network model (simple feed-forward network) on the source distribution resampled according to the outputted class weights, finetune it on limited labeled samples from the target domain, and test it against the target distribution. We provide some theoretical results for our algorithm, showing that $\algo$ can bound the error of the classifier trained on the target domain. Empirically, we compare our method with six different transfer learning methods over seven benchmark datasets to show that, on average, our approach provides the highest classification accuracy for the downstream model. 

%% file: 3methodology.tex
\section{Preliminaries}
In this section, we first introduce the notation used throughout the paper and then briefly discuss Wasserstein distance and the different settings of domain adaptation.

\textbf{Notation}
We define a domain $\gD$ as a joint distribution over the input and output space $\gX\times\gY$. In this work, we focus on the classification setting, where the output space is given by $\gY = [k] \defeq \{1, \ldots, k\}$ with $k$ being the number of output classes. In the context of representation learning, we obtain a learned representation $z = f(x)$  by applying a feature encoder $f_\theta:\gX\to\gZ$ parametrized by $\theta$, where we use $\gZ\subseteq\R^p$ to denote the feature space. Upon the feature vector $z \in f_\theta(\gX)$, we further apply a predictor $g:\gZ\to\Delta_k$, where we use $\Delta_k$ to denote the $(k-1)$-dimensional probability simplex. We use the cross-entropy loss as our objective function. More specifically, let $q_y\in \Delta_k$ be a one-hot vector with the $y$-th component being 1. The cross-entropy loss, $\lce(\cdot, \cdot)$ between the prediction $g\circ f(x)$ and the label $y$ is given by $\lce(g\circ f(x), y)\defeq \sum_{i\in[k]}q_i\log(g(f(x))_i)$. Similarly, we use $\error_\gD(h)$ to denote the 0-1 error of a classifier $h$ over the domain $\gD$, i.e., $\error_\gD(h)\defeq \E_\gD[\ell_{0-1}(h(X), Y)]$. Throughout the paper, when not specified, we shall use $\|\cdot\|$ to denote the Euclidean distance ($\ell_2$ distance). A function $h$ is called $\rho$-Lipschitz continuous if $\forall x, x'$, $\|h(x) - h(x')\| \leq\rho\|x - x'\|$. We also use $\mathbf{1}_p$ to denote a $p$-dimensional vector with all the components taking value 1. Given two matrices $A, B$ of the same size, we use $\Tr(A^\top B)$ to denote the trace of $A^\top B$, which is also the inner product of $A$ and $B$, i.e., $\Tr(A^\top B) = \sum_{ij}A_{ij}B_{ij}$.

\textbf{Wasserstein Distance}
To quantify distribution shifts between different domains, we adopt the Wasserstein distance metric~\citep{kantorovich1939}, widely used in the optimal transport literature~\citep{villani2009optimal}.
\begin{definition}[Wasserstein Distance] 
\label{def:wass}
Consider two distributions $\mu$ and $\nu$ over $\mathbb S \subseteq \R^d$. The Wasserstein distance between $\mu$ and $\nu$ is defined as 
\begin{align}
    W_1(\mu, \nu)\defeq\inf _{\gamma \in \Gamma(\mu, \nu)} \int_{\mathbb S \times \mathbb S } \|x - y\| \mathrm{~d} \gamma(x, y),
\end{align}
where $\Gamma(\mu, \nu)$ is the set of all distributions over $\mathbb S \times \mathbb S$ with marginals equal to $\mu$ and $\nu$, respectively.
\end{definition} One advantage of the Wasserstein distance over other discrepancy measures, such as the total variation, KL-divergence, or the Jensen-Shannon divergence, is that it depends on the metric being considered over $\mathbb{S}$. This property makes using it beneficial when $\mathbb{S}$ corresponds to the feature space $\gZ$ rather than the input space $\gX$.

\textbf{Domain Adaptation}
The existing generalization theory of machine learning crucially depends on the assumption that training (source) and test (target) distributions are the same~\citep{valiant1984theory}. When this assumption fails, domain adaptation (DA) focuses on adapting models trained on labeled data of a source domain to a target domain with unlabelled data. Several approaches have been proposed for DA in recent years, including domain-invariant representations~\citep{ganin2016domain,zhao2018adversarial,zhao2019domain} and self-training (i.e., pseudo-labeling)~\citep{liang2019distant,zou2018unsupervised,zou2019confidence,wang2022understanding}, among others. We will use subscripts $S$ and $T$ to denote the corresponding terms from the source and target domains to ease the notation. For example, $\gD_S$ and $\gD_T$ mean the joint distributions from the source and target domains, respectively. We shall use $\gD_S(X)$ and $\gD_T(X)$ to denote the marginal distributions over $\gX$. One recent result that uses the Wasserstein distance for domain adaptation is from~\citet{shen2018wasserstein}:
\begin{theorem}[Theorem 1~\citep{shen2018wasserstein}]
\label{thm:wbound}
Let $\gH$ be a hypothesis space where all the classifiers (score functions) are $\rho$-Lipschitz continuous. Then, for every $h\in\gH$, the following inequality holds
\begin{equation}
\error_T(h) \leq \error_S(h) + 2\rho~W_1(\gD_S(X), \gD_T(X)) + \lambda^*,
\label{equ:wbound}
\end{equation}
where $\lambda^* \defeq \argmin_{h\in\gH}\eps_S(h) + \eps_T(h)$ is the optimal joint 0-1 error that a single classifier could obtain over both domains.
\end{theorem} The above bound depends on $\lambda^*$, the optimally combined error a classifier can achieve in $\gH$. The intuition is that if the distance between the marginal distributions as measured by the Wasserstein distance is small, the better the generalization error over the target domain. 

One basic assumption in the domain adaptation literature is that both domains' input and output spaces are the same. However, this assumption could be restrictive in many applications. For example, in transfer learning, the target domain could correspond to an entirely different task so that output spaces between the source and target domains are disjoint. To bridge this gap, one new and more realistic setting, called open set domain adaptation (ODA)~\citep{panareda2017open}, has attracted increasing attention. In ODA, source and target data share only a few intersecting classes. One critical challenge in ODA is called \emph{negative transfer}~\citep{cao2018partial}, where transferring between dissimilar classes from the source domain to the target domain could instead hurt the target generalization performance, which has been confirmed both theoretically and empirically~\citep{zhao2019domain} for approaches based on learning domain-invariant representations.

\section{Method}
In this section, we present our method, \algo, in a setting, which we term as \emph{extreme label shift}, where under transfer learning, the class labels are disjoint between the source and target domains. 

\textbf{Overview of our method}
At a high level, \algo contains two stages. In the first stage, we select the best subset of classes from the source domain by solving a linear program to minimize the Wasserstein distance. In the second stage, a classifier is trained based on the reweighted source classes. Then, we fine-tune the source-trained classifier on a limited amount of data from the target domain. 

\subsection{Class-based Subset Selection via Linear Program}
\label{sec:method}
In light of the negative transfer problem when class labels are disjoint between the source and target domains, we propose to mitigate this issue by selecting a subset of classes from the source domains that are similar to data from the target domain. More specifically, let $Z = f_p(X)$ be the features we obtain after applying an encoder $f_p$, e.g., features from ResNet, to the input data. Hence, given a distribution $\gD(X)$ over $\gX$, under the features $Z = f_p(X)$, we obtain a corresponding induced distribution over $\gZ$, denoted by $\gD(Z)$. We then propose to solve the following optimization problem:
\begin{equation}
    \min_{w\in\Delta_k}\quad~W_1\left(\sum_{i\in[k]}w_i\gD_{S_i}(Z), \gD_T(Z)\right) 
\label{eq:opt}
\end{equation}
where $\gD_{S_i}(Z)$ corresponds to the marginal distribution of the features from class $i$. 
In other words, by restricting $w\in\Delta_k$ and solving for the optimal $w$ that minimizes the Wasserstein distance, we are seeking to find a mixture model of the class conditional distributions from the source domain that is closest to the target domain in the feature space. 

Before delving into how to solve the above optimization problem, discussing some of the specific design is essential. First, note that the Wasserstein distance is computed in the feature space $\gZ$, rather than the original input space $\gX$. This is important because the $\ell_2$ distance in the original input space $\gX$ does not usually reflect the similarity or difference between different data points. For example, suppose the input space $\gX$ corresponds to all the input images of a fixed size, then the $W_1$ distance on the pixel levels corresponds to the change of pixel values to transform one set of images to the other, which does not reflect the similarities of the objects in the images. Thus, it is important to choose a feature encoder $f_\theta$ so that the $\ell_2$ distance in the feature space better approximates the similarities between different input data points. Second, instead of assigning a weight for each data point, in~\eqref{eq:opt}, we assign a weight for each class. This significantly reduces the computations needed for solving~\eqref{eq:opt}. 

\textbf{An algorithm over finite samples}
In practice, the learner cannot access the underlying class-conditional distributions of the source domain. Instead, we need to estimate $W_1(\cdot, \cdot)$ based on finite samples drawn from the source domain. Based on Def.~\ref{def:wass}, we consider a source dataset $\widehat{\gD}_{S}(Z)$ with size $n$ and a target dataset $\widehat{\gD}_{T}(Z)$ of size $m$, and formulate the following linear program
\begin{align}
\label{equ:lp}
    \min_{w, P} \quad & \Tr(D^\top P) \\
    \text{s.t.}\quad & w\in\Delta_k, 0 \leq P \in \R^{n\times m}, \nonumber\\
                     & \mathbf{1}_n^\top P = \frac{1}{m} \mathbf{1}_m^\top,
                     \nonumber\\
                     & P\mathbf{1}_m = \left(\frac{w_1}{n_1}\mathbf{1}_{n_1}^\top;\cdots;\frac{w_k}{n_k}\mathbf{1}_{n_k}^\top\right)^\top\nonumber,
\end{align}
where $D\in\R^{n\times m}$ is the distance matrix between the source and target domains such that $D_{ij}\defeq \|z_i - z_j\|$ for $z_i \in \widehat{\gD}_{S}(Z)$ and $z_j \in \widehat{\gD}_{T}(Z)$. In the above optimization formulation, $P$ is the transport matrix, i.e., a joint distribution over $Z_S$ and $Z_T$. The constraint $\mathbf{1}_n^\top P = 1/m\cdot\mathbf{1}_m^\top$ ensures that the marginalization of $P$ along the first dimension equals a uniform distribution. We use $n_i, i\in [k]$ for the last constraint to denote the number of data points with class label $i$ in the source domain. Hence, this ensures that the marginalization of $P$ along the second dimension equals a re-weighted distribution of data points from the source domain, where a data point from class $i$ has weight $w_i / n_i$. Equivalently, the last constraint ensures that the weight assigned to the $i$-th class in the source domain is $w_i$.

The linear program in~(\ref{equ:lp}) implements the optimization problem in~\eqref{eq:opt} by considering the definition of Wasserstein distance (Def.~\ref{def:wass}) between the class-reweighted source and target domains. If the optimal solution $w^*$ is sparse, then only the subset of classes corresponding to the non-zero elements of $w^*$ will be considered and used during training. In practice, if $n$ or $m$ is moderately large, solving the above linear program could be computationally expensive (cubic time in $n$ and $m$). In such cases, we can apply the sinkhorn algorithm~\citep{cuturi2013sinkhorn}, which scales linearly in both $n$ and $m$.

\subsection{Transfer Learning under Extreme Label Shift}
\textbf{Pre-training}
Once we have obtained the optimal vector $w^*$, we proceed to learn a classifier $g\circ f_\theta$ from the labeled data in the source domain given by $w^*$:
\begin{equation*}
    \min_{g, f_\theta} \quad \sum_{i\in[k]: w_i^* > 0}\frac{w^*_i}{n_i}\sum_{(x, y)\in \widehat{\gD}_{S_i}} \lce(y, g(f_\theta(f_p(x))))
\end{equation*}
Note that $f_p(\cdot)$ is fixed during the above optimization since the optimal $w^*$ is obtained via $Z = f_p(X)$.

\textbf{Fine-tuning}
Because of the disjoint label classes between the source and the target domains, we cannot directly apply the learned classifier $g(\cdot)$ in the target domain. Hence, during the fine-tuning stage, on top of the learned features $f_\theta(f_p(X))$, we shall use the small amount of labeled data from the target domain to train a new classifier $g'$ as follows:
\begin{equation*}
    \min_{g'} \quad \sum_{(x, y)\in \widehat{\gD}_{T}} \lce(y, g'(f_\theta(f_p(x))))
\end{equation*}
It is worth emphasizing that the learned feature encoder $f_\theta$ will be fixed during the fine-tuning stage. Overall, our proposed approach consists of two stages. In the first stage, we select and reweight a subset of classes from the source domains by minimizing the empirical Wasserstein distance. In the second stage, we train a classifier over the reweighted source domain classes and then fine-tune the classifier on the target data.

\subsection{Theoretical Analysis}
Theorem~\ref{thm:wbound} proved that for any fixed standard classifier $h$ to be used on both domains, the target error of $h$ could be bounded by the sum of the source error, the Wasserstein distance between the two domains, and the optimal joint error $\lambda^*$. However, this does not apply to our setting since our algorithm has a fine-tuning step that generates an updated classifier tailored to the target domain. The fine-tuning step is necessary for the extreme label shift because the label spaces are disjoint. Furthermore, the proof of Theorem~\ref{thm:wbound}~\citep{shen2018wasserstein} also assumes a shared label space between $S, T$, which does not hold under extreme label shift. In such cases, it is not hard to see that $\lambda^* \geq 1$~\citep{zhao2019domain}, rendering the upper bound vacuous.

To approach the above technical difficulties, we define a lifted output space to deal with the otherwise disjoint classes. Given $\gY_S, \gY_T$ to represent the class set of $S, T$, let $\gY_{S,T} = \gY_S \cup \gY_T$ be the union of the two sets as the lifted output space bridging the two domains under a common label set. Under this construction, it is clear that any probabilistic classifier over $\gY_S$ or $\gY_T$ is still a probabilistic classifier over $\gY_{S,T}$. For a probabilistic classifier $h:\gX\to\Delta_k$, we first define the so-called induced classifier from $h$ as follows.
\begin{definition}[Induced classifier]
    Let $h:\gX\to \Delta_k$ be a probabilistic classifier. Define an induced classifier $\hat{h}:\gX\to [k]$ as follows: $\forall i\in[k], \hat{h}(X) = i$ with probability $h(X)_i$. 
\end{definition}
The induced classifier $\hat{h}$ is a randomized classifier that outputs the class label according to the probability vector given by $h(X)$. This is different from the deterministic classifier that always outputs $\argmax_{i\in [k]}h(X)_i$. The following proposition gives a closed-form characterization of the 0-1 classification error of any induced classifier in terms of the probabilistic classifier. 
\begin{proposition}
    Let $h:\gX\to \Delta_k$ be a probabilistic classifier and $\hat{h}$ its induced classifier. Then, the expected error of the induced classifier $\error_\gD(\hat{h}) = \frac{1}{2}\E\left[\|h(X) - Y\|_1\right]$, where $Y\in\{0,1\}^k$ is a $k$-dimensional one-hot vector representing the ground-truth label.
\end{proposition}
Based on the above characterization, the next lemma bounds the error difference of the same probabilistic classifier $h$ under two domains $\gD_S$ and $\gD_T$ by their Wasserstein distance:
\begin{proposition}
\label{prop:diff}
    Let $h:\gX\to \Delta_k$ be a probabilistic classifier that is $\rho$-Lipschitz continuous under the input norm $\ell_2$ and output norm $\ell_1$, i.e., $\forall x, x'\in\gX$, $\|h(x) - h(x')\|_1\leq \rho\|x - x'\|_2$. Then, for the corresponding induced classifier $\hat{h}$, we have $|\error_S(\hat{h}) - \error_T(\hat{h})| \leq \max\{\rho, 1\}\cdot W_1(\gD_S, \gD_T)$.
\end{proposition}
We would like to emphasize that the distributions $\gD_S$ and $\gD_T$ in~\Cref{prop:diff} are the joint distributions from source and target domains (not the marginal ones). The following lemma then further decomposes the Wasserstein distance between the joint distributions by the sum of the Wasserstein distance between the marginal distributions and the conditional distributions.
\begin{lemma}
\label{lemma:decomp}
    For any two joint distributions $\gD_S$ and $\gD_T$ over $\gZ\times\gY$, we have
    \begin{align}
        &W_1(\gD_S, \gD_T) \leq W_1(\gD_S(Z), \gD_T(Z)) \nonumber \\
        &+ \min\Big\{\E_{\gD_S(Z)}[W_1(\gD_S(Y\mid Z), \\ 
        &\phantom{+ \min\Big\{}\gD_T(Y\mid Z)], \nonumber \\
        &\phantom{+ \min\Big\{}\E_{\gD_T(Z)}[W_1(\gD_S(Y\mid Z), \nonumber \\
        &\phantom{+ \min\Big\{}\gD_T(Y\mid Z)]\Big\}. \nonumber
    \end{align}
\end{lemma}
Now, combine all the results above and note that~\Cref{prop:diff} works for any pair of joint distributions $\gD_S$ and $\gD_T$, including the reweighted one, we have the following upper bound:
\begin{align*}
\error_T(\hat{h}) \leq& \min_{w\in\Delta_k} \error_{S(w)}(\hat{h}) + \\
&\max\{\rho, 1\}\cdot \Big(W_1(\gD_{S(w)}(Z), \gD_T(Z)) \\
&+ \min\big\{\E_{\gD_{S(w)}(Z)}[W_1(\gD_{S}(Y\mid Z), \\
&\phantom{+ \min\big\{}\gD_T(Y\mid Z)], \\
&\phantom{+ \min\big\{}\E_{\gD_T(Z)}[W_1(\gD_{S}(Y\mid Z), \\
&\phantom{+ \min\big\{}\gD_T(Y\mid Z)]\big\}\Big)
\end{align*}
where we use $S(w)$ to denote the reweighted source domain by classes. Note that the conditional distribution $\gD_S(Y\mid Z)$ is invariant under the reweighting, so to minimize the upper bound, it suffices for us to minimize the first and second terms in the upper bound, which inspires the design of our method in~\Cref{sec:method} by finding the reweighting vector $w$ and classifier $h$ that jointly minimize the first two terms.

The last part of our method transfers a classifier $h$ from the reweighted source domain $S(w)$ to the target domain $T$. However, the previous bound assumes the same classifier $h$ for both domains. The following theorem takes into account the difference in classifiers.  ~\citep{mousavi2020minimax} proved the following theorem:
\begin{theorem}
\label{theorum:transfer}
Given a classifier $h(\Omega, x) \coloneqq h(V, x) = V\phi(Wx)$, define two classifiers $V_{S}(Wx), V_{T}(Wx)$ as the pre-trained and fine-tuned models where only the final layer is changed. The transfer distance between the two models can be represented as:
\begin{align*}
    \Omega(V_{S}-V_{T}) = ||\Tilde{\sum^{\frac{1}{2}}_{T}}(V_{S}-V_{T})^T||_F
\end{align*}
\end{theorem} 
We can extend this theorem to provide a bound between a pre-trained and finetuned classifier as follows:
\begin{lemma}
\label{lemma:transfer}
Given two classifiers $h,h'$ that only differ by their final layer, we can bound their error as following:
\begin{align*}
   \epsilon_T(h') \leq \epsilon_S(h) + \rho W_1 + \alpha \beta \sigma_{\max} (V_{S}-V_{T})
\end{align*}
\end{lemma}

where  $\sigma_{\max}$ is the largest singular value over $(V_{S}-V_{T})$, $\alpha$ is the Lipchitz constant that bounds the softmax function, and $\beta$ is a bound on the L2 norm of the feature space. 
This results in the following upper bound:
\begin{align*}
\error_T(\hat{h'}) \leq& \min_{w\in\Delta_k} \error_{S(w)}(\hat{h}) \\ 
&+ \max\{\rho, 1\}\cdot \Big(W_1(\gD_{S(w)}(Z), \gD_T(Z)) \\
&+ \min\big\{\E_{\gD_{S(w)}(Z)}[W_1(\gD_{S}(Y\mid Z), \\
&\phantom{+ \min\big\{}\gD_T(Y\mid Z)], \\
&\phantom{+ \min\big\{}\E_{\gD_T(Z)}[W_1(\gD_{S}(Y\mid Z), \\
&\phantom{+ \min\big\{}\gD_T(Y\mid Z)]\big\}\Big) \\
&+ \alpha \beta \sigma_{\max} (V_{S(W)}-V_{T})
\end{align*}
where $V_{S(W)}$ is the last layer of the model trained on the reweighted source distribution and $V_{T}$ is the last layer of that same model that has been fine-tuned on the target domain.

%% file: 4results.tex
\section{Experiments}
\label{sec:res}
Previously, we provided a theoretical analysis upper-bounding WaSS's error. We now empirically evaluate WaSS's efficacy in selecting the optimal subset of training classes. We compared \Sys against five different baselines on four separate datasets. For the brevity of the paper, we will briefly detail the setting for each experiment and leave additional details, including details about the datasets, baselines, and hyperparameter selection, in \autoref{sec:app:add}. For each experiment, we created a source and target domain by splitting the classes in each dataset. The class set was divided so the target domain contained $3$ classes while the rest of the classes represented the source domain. Depending on the label shift setting, the source domain could additionally include a subset of the test classes. We plan to release our implementation of \Sys upon publication.

% Particularly, we investigate the following research questions (RQs):
% \begin{itemize}
%     \item \textbf{RQ1:} How much does WaSS's weighting of the source classes quantitatively impact the transferability between the source and target domain?
%     \item \textbf{RQ2:} How does WaSS weighting of the source domain's classes qualitatively compare to the target domain class set?
%     \item \textbf{RQ3:} Is there a correlation between WaSS's performance and the number of dimensions of the embedding space?
%  \end{itemize}

% To answer these questions, we 

\subsection{Quantitative Performance}
\label{sec:res:quant}
% Our first research question (RQ1) seeks to
We evaluate WaSS's impacts on the transfer distance between the source and target domain; however, measuring transfer distance is complex \cite{jiang2022transferability}. Instead, we use the accuracy of a feed-forward neural network on the target distribution as the proxy for calculating transfer distance. We conduct our experiment on two label shift scenarios: disjoint (DDA) and open-set (ODA).

\input{tables/table1}
\input{figures/figure4}

% As pointed out in \autoref{sec:rel_work}, in contrast to previous work, WaSS is inherently robust to the degree of label shift between domains as WaSS recovers the optimal class weighting using Wasserstein distance within the input space. To demonstrate this, i

\textbf{Disjoint Set (DDA):} In Table \ref{tab:results_table}, we present the accuracies for five different methods within the Disjoint Set Domain Adaptation (DDA) setting - a complete absence of class overlap between the source and target domains. For each dataset, we executed every method across three distinct source/target domain class splits and reported the average accuracy over ten iterations. For all but two experiments, WaSS's class weighting on the source domain best facilitates the transfer of the downstream classifier to the target distribution, resulting in an average $1-2$ absolute percentage point increase over the next-best selection strategy. In both instances where WaSS is outperformed, a uniform weighting over the source classes (ALL) provided a better transfer between domains. We theorize that WaSS's dependency on a finite number of input samples to calculate the optimal subset can result in errors within the weighting. For situations where the optimal class selection is close to the source class distribution, the errors induced by the finite-sample effect allows ALL to provide a more accurate class selection than WaSS.

We further validated our experiments by conducting a large-scale test on Fashion-MNIST and CIFAR-10 sampling $40$ different source/target domain class splits. We present the results in Figure \ref{fig:boxplot} where, in both datasets, WaSS (statistically significantly \footnote{p < 1e-4; see \autoref{sec:stats} for details of statistical testing}) outperforms all the other baselines  - generating the highest mean accuracy on both datasets. However, on CIFAR-10, All and PADA achieve higher accuracies than WaSS for select source/target domain class splits, possibly due to the finite-sample effect, as previously discussed.

PADA and OSS represent previous work within the domain adaptation and label shift area - while competitive, these methods do not outperform WaSS. PADA and OSS try to identify similar classes between domains instead of recovering optimal subsets - making them dependent on a shared label set existing between domains. To better compare with these methods, we next evaluate the performance of WaSS within an ODA setting where there exists an overlap of each domain's class set.

\input{figures/figure9}

% We have shown that WaSS can significantly outperform all the other baselines when the source and target class sets are disjoint. However, 

\textbf{Open Set (ODA):}  DDA violates the assumptions made by previous work within the transfer learning and label shift area. To assess how WaSS compares against these methods, we consider an open-set domain adaption (ODA) setting where the source domain contains a subset of the classes within the target domain \ref{sec:rel_work}. 

% where a subset of classes exists in both the source and target class sets. 

% We create source and target domains by splitting datasets based on their label sets. For each split, we enforce that two of the three classes in the target domain are present within the source domain's class set. 

In Figure \ref{fig:oda:acc}, we provide the accuracy of the downstream classifier when evaluated against the target domain. Additionally, Figure \ref{fig:oda:dist} reports the TV distance calculated based on the weights of classes intersecting between domains. In both figures, WaSS performs better than OSS over all datasets, indicating that the resulting downstream classifier can better generalize to the target domain. This can be attributed to WaSS weighing the source classes with respect to the distance between the source and target domain, while OSS only selects the classes that occur within both domains. As a result, WaSS chooses classes within the source domain that allow the downstream classifier to better generalize to the classes that are not shared between the source and target domain.

\input{figures/figure1}

\subsection{Qualitative Performance}
% Our second research question (RQ2) looks for qualitative results that show how WaSS's weighting brings the source domain closer to the target domain.

We now qualitatively show how WaSS's weighting can reduce the distance between the source and target domains. To visualize `closeness,' we utilize pie charts that describe the transformation of the source class distribution before and after WaSS's reweighting. Beneath each figure, we also provide the target domain class set to show the increase in similarity between the target domain and selected source domain classes. 
\input{figures/figure5}

\textbf{Disjoint Set (DDA):} To visualize closeness in DDA, we show that the classes selected by WaSS are semantically close to the classes in the test set as there is no class overlap between the two domains. Figure \ref{fig:pie:openset} exhibits WaSS's class selection for a sample train/test split in Fashion MNIST and Cifar-10. The source domain has a uniform weight across all its classes (the inner circle) for both datasets. WaSS weights the source class set (outer circle) to select a subset of the classes and remove classes that hinder domain transfer. For example, the classes in the target domain for Fashion MNIST are $\{\text{T-shirt/top, Trouser, Pullover}\}$. WaSS narrowed the source domain class set from $\{\text{Dress, Coat, Sandal, Shirt, Sneaker, Bag, Ankle boot}\}$ to $\{ \text{Dress, Shirt, Coat}, \}$ getting rid of non-clothing classes like bag and ankle boots. In Appendix \ref{sec:pie_charts}, we provide an additional pie chart visualizing WaSS selection for PACS within the disjoint setting. PACS differs from the previous two datasets by featuring a non-uniform weighting for each class within the source dataset. WaSS still selects a subset of classes semantically similar to the target domain.
\input{figures/figure8}

\textbf{Open Set (ODA):} In ODA, where a subset of the label spaces is shared between domains, `closeness' is related to identifying the class overlap. However, `closeness' also measures how non-overlapping classes selected in the source domain relate to the non-overlapping class in the target domain. For example, in figure \ref{fig:pie:openset}, the result for WaSS and OSS on a single test/train split within Fashion-MNIST is visualized. The target domain class set is $\{ \text{Ankle boot, T-shirt/top, Trousers}\}$ with $\{ \text{Trousers}\}$ being the non-overlapping class. Both OSS and WaSS correctly identify $\{ \text{T-shirt/top, Trousers}\}$ as important classes for transfer. However, only WaSS matches the clothing-heavy target domain class set by weighting to all clothing classes within Fashion-MNIST $\{\text{Dress, Pullover, Shirt, Coat, Sneakers}\}$ while dropping irrelevant non-clothing classes $\{\text{Bag}\}$. Conversely, because OSS selects only overlapping classes between domains - ignoring the additional source domain classes - the downstream classifier associated with WaSS outperforms the downstream classifier associated with OSS for every open-set scenario (Figure \ref{fig:oda:acc}) as it can better generalize to the target domain.

%% file: tables/table1.tex
\begin{table}
  \caption{Accuracy of the downstream classifier trained on source distribution weighted by classes selected according to each baselines method in total class disjointness on MNIST-Fashion (M-Fash), Cifar 10 (C-10), PACS (P), Cifar 100 (C-100). Best performing results are bolded. We use a paired $t$-test to show our results' statistical significance (p < 0.05).}
\label{tab:results_table}
\small
\resizebox{\columnwidth}{!}{
  \centering
  \begin{tabular}{llllllll}
%    \toprule
%    \cmidrule(r){1-2}
    Dataset     & Test Class  & All  & PADA & RND & MN & OSS &WASS  \\
    \midrule
M-Fash & $[0,1,2]$        &  74.92 &71.88 &57.63 &75.36& 39.68 &\textbf{78.76} \\
M-Fash & $[3,4,5]$        &  36.10 &37.33 &34.65 &35.00& 38.53 &\textbf{40.34} \\
M-Fash & $[6,7,8]$        &  63.15 &68.28 &50.75 &56.61& 63.55  &\textbf{68.63} \\
M-Fash & $[9,0,1]$        &  47.80 & 41.05 & 45.89 & 40.72& 49.78 & \textbf{54.64} \\
\hline
C-10   & $[0,1,2]$        &  70.17 &70.76 &48.56 &69.51& 68.27 &\textbf{73.19} \\
C-10   & $[3,4,5]$        &  86.22 &84.51 &72.50 &82.03& 79.82 &\textbf{89.77} \\
C-10   & $[6,7,8]$        &  77.77 &70.88 &61.15& 67.51& 80.09 & \textbf{82.53} \\
C-10   & $[9,0,1]$        &  77.26 &75.25 &65.16 &76.67& 53.29 &\textbf{77.85} \\
\hline
PACS      & $[0,1,2]$        &  35.00 &32.01 &34.65 &33.35& 43.20  &\textbf{38.43} \\
PACS      & $[3,4,5]$        &  \textbf{64.63} &45.68 &48.85 &33.34 & 56.60 &63.10 \\
PACS      & $[1,2,6]$        &  39.66 &33.75 &30.16 &32.98& 34.27  &\textbf{41.77} \\
PACS      & $[0,3,5]$        &  55.91 &41.95 &37.75 &33.34&  49.48&\textbf{55.95} \\
\hline
C-100  & $[72, 4, 95]$    &  57.62 & 27.83 & 56.83 & 55.67& 56.82  &\textbf{58.87} \\
C-100  & $[73, 32, 67]$   &  68.44 & 68.30 & 62.33 & 63.17&  41.66 &\textbf{70.30} \\
C-100  & $[92, 70, 82]$   &  74.90 &80.10 & 73.00 &60.50& 60.60 &\textbf{82.70} \\
C-100  & $[16, 61, 9]$    &  \textbf{93.66} &90.88 &89.66 & 92.66& 64.88 &92.56 \\
% \hline
% Average  &  & 63.95 & 58.78 & 54.35 & 56.78 & 55.03 & \textbf{66.84} \\
    \bottomrule
  \end{tabular}
 }

\end{table}

%% file: figures/figure4.tex
\begin{figure*}[!tb]
\centering
	\begin{subfigure}[]{0.49\textwidth}
		\includegraphics[width=\textwidth]{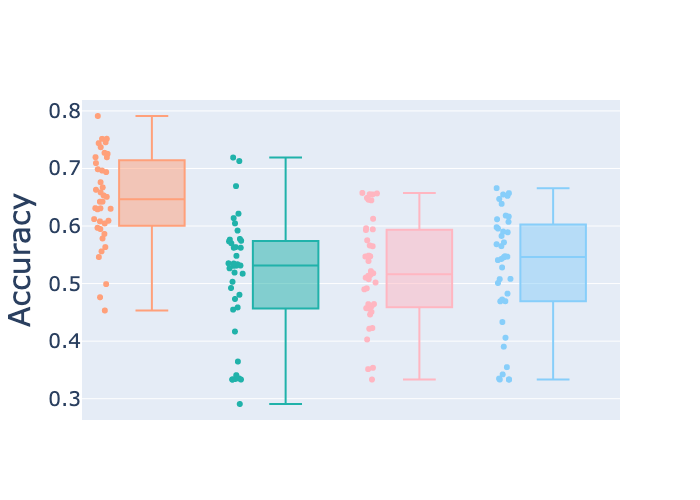}
		\caption{Fashion-MNIST}
		\label{fig:mnist40}
	\end{subfigure}
	\begin{subfigure}[]{0.49\textwidth}
		\includegraphics[width=\textwidth]{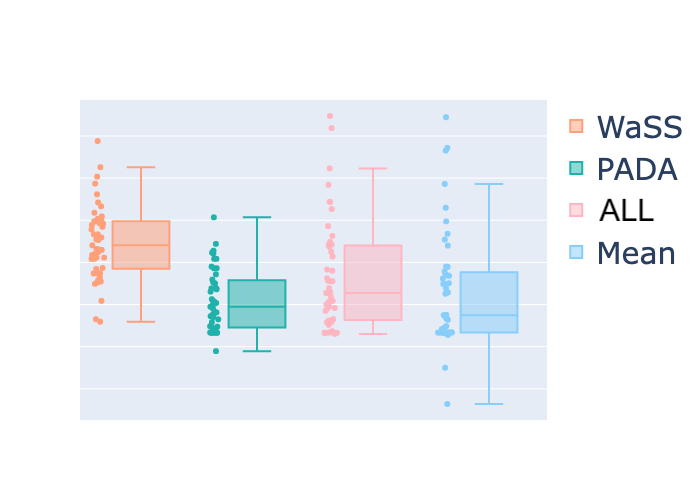}
		\caption{CIFAR-10}
		\label{fig:cifar40}
	\end{subfigure}
    \caption{}
    \label{fig:boxplot}
\end{figure*}

%% file: figures/figure9.tex
\begin{figure}[!h]
\centering
	\begin{subfigure}[]{0.5\textwidth}
		\includegraphics[width=\textwidth]{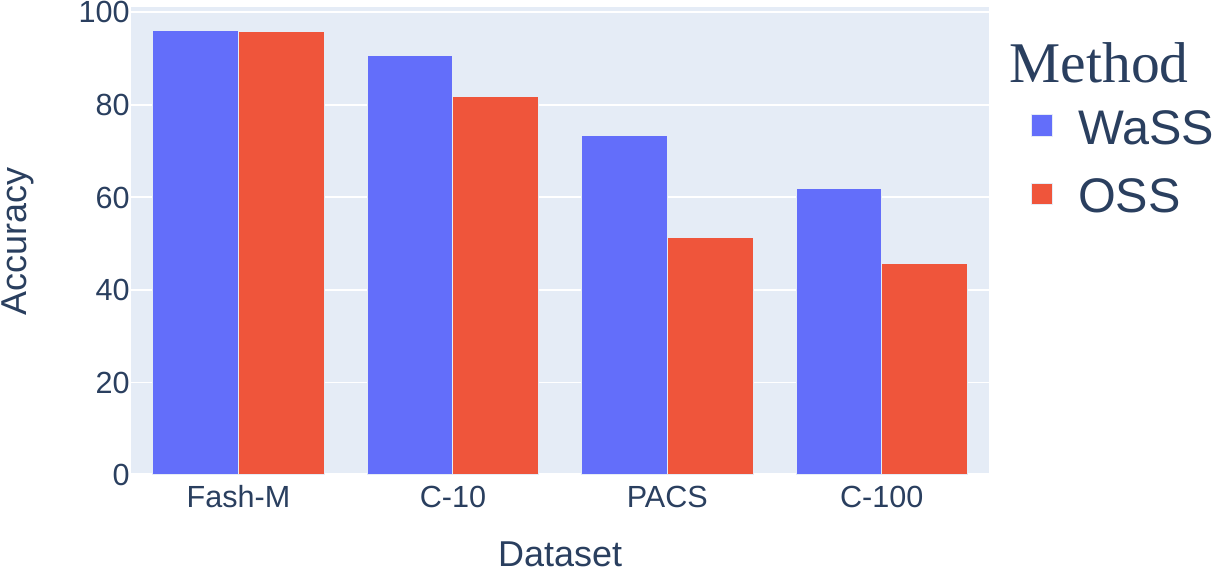}
		\caption{Accuracy}
		\label{fig:oda:acc}
	\end{subfigure}
	\begin{subfigure}[]{0.49\textwidth}
		\includegraphics[width=\textwidth]{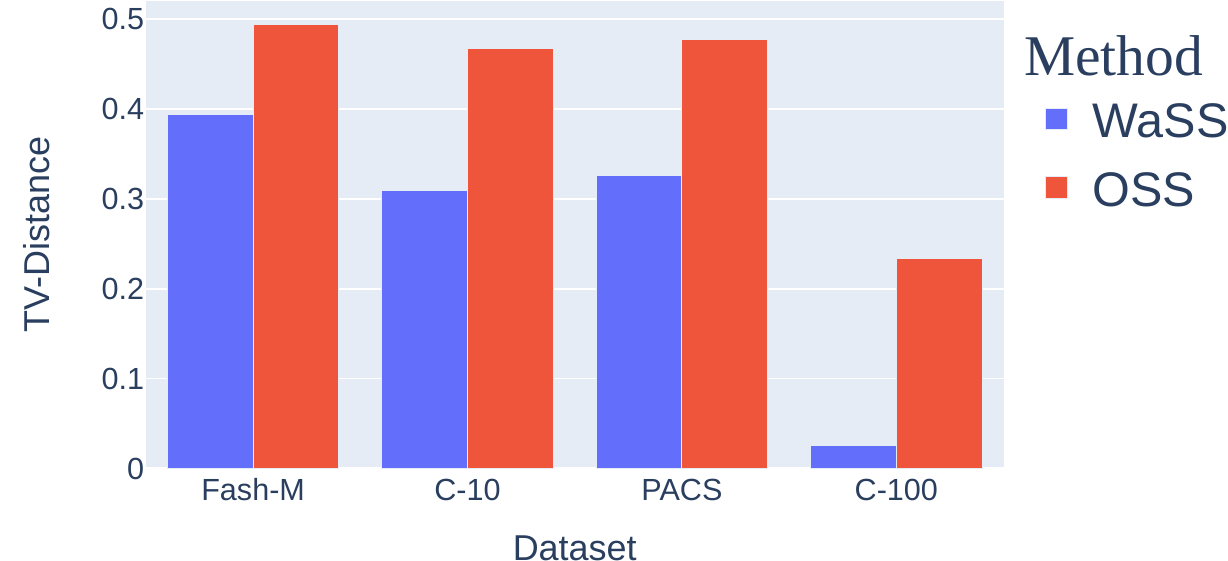}
		\caption{TV-Distance}
		\label{fig:oda:dist}
	\end{subfigure}
    \caption{Accuracy of the downstream classifier on the target domain weighted by class selection methods: OSS and ALL within an open-set setting. TV Distance of two different class selection methods based on overlapping classes between domains.}
    \label{fig:open-set}
\end{figure}

%% file: figures/figure1.tex
\begin{figure}[!tb]
\centering
	\begin{subfigure}[]{0.33\textwidth}
		\includegraphics[width=\textwidth]{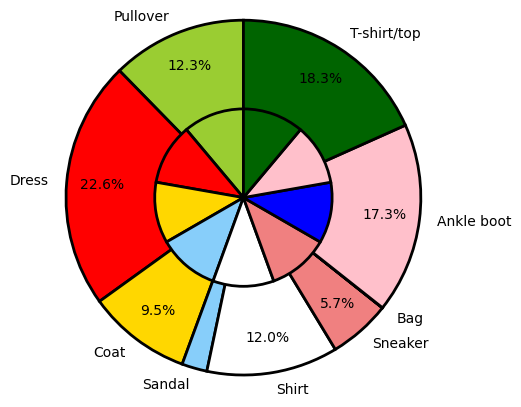}
		\caption{F-MINST}
		\label{fig:histo:mnist}
	\end{subfigure}
	\begin{subfigure}[]{0.29\textwidth}
		\includegraphics[width=\textwidth]{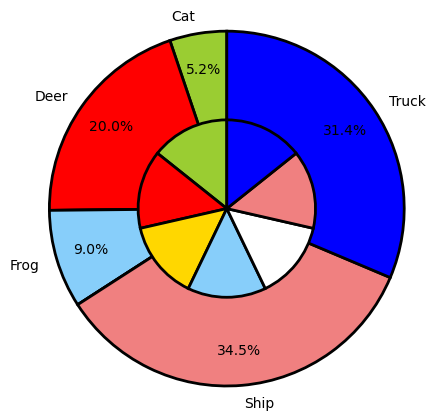}
		\caption{CIFAR-10}
		\label{fig:histo:cifar10}
	\end{subfigure}
    \caption{Class distributions before (inner circle) and after (outer circle) applying our method to arbitrarily selected test classes. The same color corresponds to the same class (e.g. Deer changes from 14\% to 20\%). Test Classes for F-MNIST are $\{\text{T-shirt/top, Trouser, Pullover}\}$ and for Cifar-10 are $\{\text{Airplane, Automobile, Bird}\}$.}
    \label{fig:pie:disjoint}
\end{figure}

%% file: figures/figure5.tex
\begin{figure}[!tb]
\centering
	\begin{subfigure}[]{0.34\textwidth}
		\includegraphics[width=\textwidth]{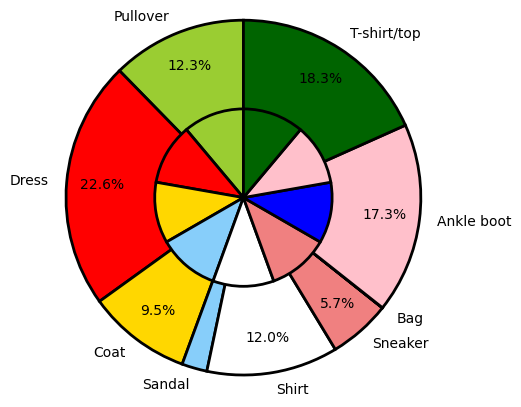}
		\caption{WaSS}
		\label{fig:pie:openset:mnist:wass}
	\end{subfigure}
	\begin{subfigure}[]{0.25\textwidth}
		\includegraphics[width=\textwidth]{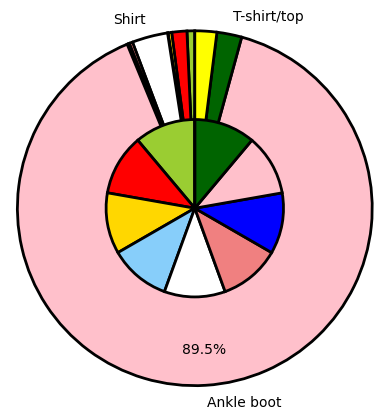}
		\caption{OSS}
		\label{fig:pie:openset:mnist:oss}
	\end{subfigure}
    \caption{Class distributions of test set (inner circle) and class distribution of training set (outer circle) weighted by class selection methods: WaSS and OSS. The same color corresponds to the same class. Test Class: $\{\text{Ankle boot, T-shirt/top, Trousers}\}$}
    \label{fig:pie:openset}
\end{figure}

%% file: figures/figure8.tex
\begin{figure}[!tb]
\centering
	\begin{subfigure}[]{0.5\textwidth}
		\includegraphics[width=\textwidth]{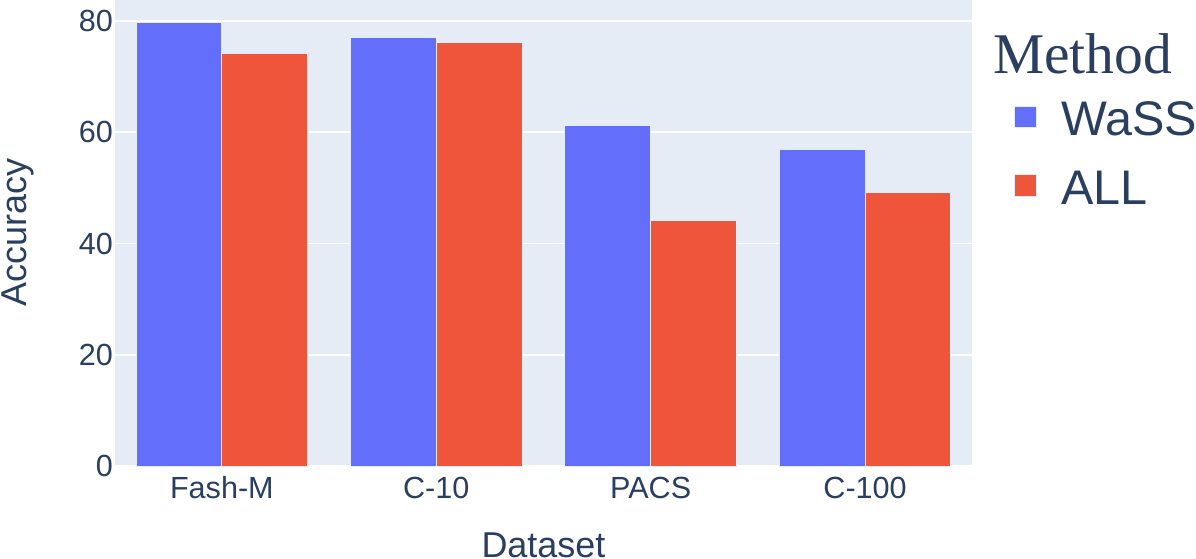}
		\caption{64 Features}
		\label{fig:feat:64}
	\end{subfigure}
	\begin{subfigure}[]{0.49\textwidth}
		\includegraphics[width=\textwidth]{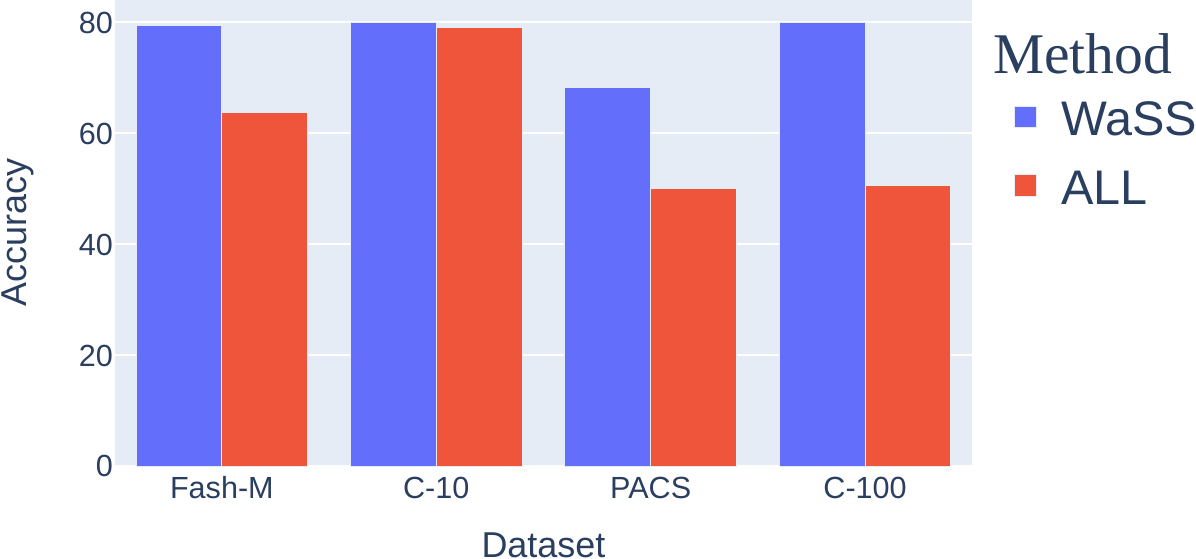}
		\caption{256 Features}
		\label{fig:feat:256}
	\end{subfigure}
    \caption{Accuracy of the downstream classifier on the target domain weighted by class selection methods: WaSS and ALL under two different embedding dimensions.}
    \label{fig:feature-sz}
\end{figure}

%% file: 2related.tex
\section{Related Works}
\label{sec:rel_work} 

\textbf{Domain Adaption}
Based on the degree of separation between domains, previous work can be classified into three different settings: Closed Domain Adaptation (CDA), Partial Domain Adaptation (PDA), and Open Set Domain Adaptation (ODA). In CDA ~\citep{saenko2010adapting,gong2012geodesic, pan2010domain, tzeng2014deep, long2015learning, ganin2016domain}, the class set between source and target domains is assumed to be the same, so methods focus on mitigating the distributional discrepancy within the input space. PDA ~\citep{cao2018partial, cao2018partial2, zhang2018importance, li2020deep} assumes the source class set is a superset of the target class set; previous methods filter out source classes to match the class set between the domains. In ODA ~\citep{panareda2017open, saito2018open, liu2019separate}, an intersection exists between the source and target domain class sets. Previous work trains classifiers to identify intersecting class sets -- marking other classes as unknown. Our method differs from previous work by using the input distribution to measure the similarity between label sets. 

\textbf{Wasserstein Distance}
Compared to other distance metrics ~\citep{kullback1951information, lin1991divergence, massey1951kolmogorov}, Wasserstein distance is symmetric and provides a smooth representation of distance even between distributions of non-overlapping class sets. Research in transfer learning has frequently utilized Wasserstein distance; ~\citep{courty2017joint, flamary2016optimal, damodaran2018deepjdot, lee2019sliced, shen2018wasserstein} propose methods that create an optimal match between samples in the source and target domain by jointly optimizing the Wasserstein distance over dataset properties like class regularity and feature distribution. Our work differs from previous work using Wasserstein distance for domain adaptation by considering class-level weighting agnostic to any label shift occurring. 

% ~\citep{redko2019optimal} uses Wasserstein distance to address label shift across multi-source domain adaptation.

\textbf{Transferability Measurement}
Transferability estimation seeks to quantitatively estimate the ease of transferring knowledge from one classification task to another. Previous work ~\citep{ben2003exploiting,ben2006analysis,dasu2006information, mansour2009domain} has presented several estimation techniques for capturing model performance on the target domain. Our method differs from previous research by defining Wasserstein distance as a proxy for transferability estimation and selecting source classes that optimally reduce that distance.

%% file: 5conclusion.tex
\section{Conclusion}
While large-scale pre-training has benefited many deep learning methods, ensuring this pre-training data is appropriate for the task at hand is critical. To this end, we introduce a new method, WaSS, that identifies the subset of source domain classes that optimally reduce the transfer distance between source and target domains. Conventional wisdom may dictate it is better to use all available data, but this can be problematic due to negative transfer, especially for smaller datasets. Our method uses an efficient linear programming method employing Wasserstein distance as a proxy for transferability, which allows us to outperform all baselines and provide a more effective source class weighting.

%In this project, we investigate the identification of efficient subsets for transfer learning. We run numerous empirical studies on the correlation between Wasserstein distance and finetuned accuracy, but results are unclear. It shows exciting results, but further investigation is necessary to ensure consistent performance. Based on this correlation, we implement two algorithms to efficiently identify low Wasserstein distance subsets that will be effective for transfer learning on the target domain from the source domain.

%In summary, we devise a method, super subset selection ($S^3$), to find the most useful training classes which could achieve a high accuracy for testing. We use Wasserstein distance, prove the upper bound for our method, and apply them into our two algorithms. Using our algorithms, we can efficiently select a subset for training an effective transferable model. 

%In the future, we plan to investigate an extension of our algorithm to data points. This would allow more fine-grained control over the training dataset. Additionally, this would also be quite interesting as an anomaly detection algorithm. We may also investigate our subset selection algorithm for calculating the bound. However, in practice we found that it does not cause a significant impact on the search algorithm.

\acknowledgments{We thank Han Zhao for helpful discussions and assistance with proving the generalization bounds. His insights and expertise in statistical learning theory were invaluable to this work.}

%% file: 6appendix.tex
\section{Additional Details About Experimental Setup}
\label{sec:app:add}
\paragraph{Datasets}
We evaluate the performance of our method in Section \ref{sec:res} on four datasets: Fashion-MNIST \citep{xiao2017/online}, CIFAR-10 \citep{Krizhevsky2009LearningML}, PACS \citep{zhou2020deep}, and CIFAR-100 \citep{krizhevsky2009learning}. \textbf{Fashion-MNIST} is the simplest dataset we consider containing $60,000$ train and $10,000$ test grayscale images of ten different clothing items. Within the dataset, only a handful of features (shape, size) differentiate each class. \textbf{CIFAR-10} is a more complex dataset consisting of $10$ different classes with a train/test split of $50,000/10,000$  colored images, respectively. Adding color allows for a richer feature set per class within the dataset. \textbf{PACS} is a dataset that simulates the domain adaptation problem; it contains 4 domains, each with 7 classes. Furthermore, PACS does not provide a constant number of images per class as to emulate real-world scenarios. \textbf{CIFAR-100} is a larger extension of CIFAR-10, containing $100$ classes with a training/test split of $50,000/10,000$ colored images, respectively. CIFAR-100 provides course-grained labels that group its $100$ classes into $20$ domains. PACS and CIFAR-100 represent the noisy datasets machine learning often encounters: with PACS modeling class imbalance and domain shift and CIFAR-100 modeling diverse and numerous class labels.

We sample three classes for each dataset to serve as the target domain and use the remaining classes as the source domain. We show $4$ different source/target splits to account for inter-dataset variability in \autoref{tab:results_table}. For CIFAR-100, we maintain the dataset structure by sampling the test domain from a single superclass and using the other superclasses as the source domain. Similarly, for PACS, the three classes composing the target domain are selected from one of the distributions within the dataset while the other $3$ distributions are used for training. 

\paragraph{Implementation} All our experiments are implemented using PyTorch \citep{NEURIPS2019_9015}.
We use a ResNet-50 pre-trained on ImageNet \citep{5206848} to create a $2048$ dimensional feature vector for each image in the dataset. Within the embedding space, we calculate the Wasserstein distance between different images. \Sys aims to find the optimal weighting of classes within the source dataset such that the Wasserstein Distance between the resulting mixture of source classes and the target domain is optimally close. We use POT \citep{flamary2021pot} to calculate the Wasserstein distance between the source and target domains. We utilize SinkHorn \citep{cuturi2013sinkhorn} to increase the efficiency of our algorithm, especially when presented with larger datasets such at CIFAR-100 and PACS. Once we solve for the optimal mixture, we train a $3$-layer fully connected neural network model on the ResNet embedded training dataset sampling according to the resulting class weights. To ensure a fair comparison between the different baselines, we sample a fixed number of images from the source dataset - preventing a larger selection of classes from resulting in a larger dataset for the model to train on. Given there is a class imbalance within PACS, we randomly oversample in the minority class to rebalance the dataset ensuring an even number of images per class.

To transfer to the target domain, we freeze  all the parameters of our downstream classifier except for its final layer. We then finetune the network on a small set of labeled images from the target domain -e.g. $100$ images per test class. Given the size of the dataset, we use $K$-folds to create a validation set and identify validation loss. This allos us to utilize early stopping when training the downstream classifier, stopping when the validation loss stagnates for over $5$ epochs. To test the downstream classifier, we evaluate its performance against the rest of the samples within the target dataset. We repeat this procedure $10$ times and report the average accuracy.  We also use cross entropy as our loss function, Adam as our optimizer, and ReduceLROnPlateau scheduler to optimally set the learning rate starting at $1e4$. For each iteration of the same experiment, we change the seed to ensure diversity within the initial parameters of the downstream classifier. 

\paragraph{Baseline Models}
We compare WaSS against four different baseline methods for subclass selection.

\textbf{ALL} is the naive baseline that selects all classes within the source domain. This is the most straightforward subset selection method but also the most prone to negative transfer. \textbf{PADA} or Partial Adversarial Domain Adaptation \cite{cao2018partial} is a domain adversarial neural network that provides class-level weights to the source domain. PADA elicits these weights by training a classifier on the source domain and then averaging over the output probabilities on the target domain. PADA asserts that source classes closest to the target domain have a higher probability in the output. \textbf{RAND} is a random sampling of all the possible subsets within the source domain. For each $10$ iteration we run on the same source and target classes, we resample and select a new subset. \textbf{MN} selects the closest source classes to the target domain by computing the mean embedding for each class within the source dataset and labels the target dataset used for finetuning. The source classes with embeddings closest to the target are picked as the optimal subset. \textbf{OSS} or Open Set Domain Adaptation by Backpropagation \cite{saito2018open} uses a feature generator and a discriminator to identify the shared classes between the source and target domain. \cite{saito2018open} weakly trains a discriminator to identify any target sample as an unknown with probability $p$. The feature discriminator tries to increase the error of the discriminator by either matching the source and target distribution for large $p$ or making the source and target distribution separate for small $p$. Within the paper, Saito et al. use a $p$ of $0.5$ such that the feature generator only matches the samples from the overlapping classes within the target domain to the source domain.

\section{Limitations}
\Sys identifies the optimal subset of classes within the source distribution to facilitate transfer between datasets. However, as we mentioned within our experimental section \ref{sec:res}, our method does not always outperform the baselines due to a finite sample effect. As a result, our method may not result in the optimal subset if not enough number of samples are provided from the target distribution. 

Another limitation of \Sys is the number of datasets we experimented with. As previously stated, our dataset selection sampled datasets of increasing noise. However, there exist more datasets that are used within the literature, and to better understand the performance of our method, we could experiment with these datasets. In Section \ref{app:addExp}, we do experiment with $3$ other datasets that are more complicated and diverse than the $4$ datasets we previously introduced. In these experiments we show that \Sys can provides a better accuracy performance than the other baselines it is compared against. 

Finally, our method uses a pre-trained deep neural network to embed each image in the dataset. These embeddings provide a semantically useful similarity measure that we then utilize to identify the optimal subset. However, a pre-trained deep neural network may not always provide a meaningful embedding to images within a dataset. This can skew our method and result in a subset not fully representing the similarity between domains. Future work can look into the effects the neural network used to embed the images may have on the performance of \Sys. 

\section{Additional Experiments}
\label{app:addExp}
\input{tables/table4}
In section \ref{sec:res}, we evaluate against four different datasets to show the benefits brought on by \Sys empirically. However, there are limitations in the datasets we evaluate on including their scale and diversity of images. As a result, we consider three additional datasets to evaluate \Sys in a Disjoint Set Domain Adaptation (DDA) setting: Office-31, Office-Home, VizDa. The Office-31 dataset comprises $31$ object categories from Amazon, DSLR, and Webcam domains, showcasing common office items. Amazon provides 2817 clean-background images, DSLR offers $498$ high-resolution captures, and Webcam presents 795 images with notable noise and color artifacts. The dataset aims to represent diverse office settings with varying image characteristics.  Office-Home is a  dataset that encompasses four domains, each comprising $65$ categories. The domains include Art, featuring artistic images like sketches and paintings; Clipart, a collection of clipart images; Product, displaying objects without backgrounds; and Real-World, showcasing objects captured with a regular camera. With a total of $15500$ images, the dataset maintains an average of approximately $70$ images per class, with some classes reaching a maximum of $99$ images. The 2017 Visual Domain Adaptation (VisDA) dataset features $3$ different domains each with $12$ categories for a total of $280000$ images for object classification. The training images are synthetic images generated of different objects in various scenarios, while the validation and test are realistic images of different objects sourced from the MSCOCO dataset.

\input{figures/figure7}

We repeat the experiments we conducted within section \ref{sec:res:quant} for each dataset - randomly sampling three classes within the class set to represent the target domain and letting the rest of the classes represent the source domain. Similar to PACS, three of these datasets are used to emulate distribution shifts for the same set of classes. Similar to PACS, to encourage \Sys to find similarity between domains we changed the train/test split. Instead of sampling three classes for test and using the rest of the classes for train, we initially selected a single distribution within the dataset from which we sampled three classes. These three classes represented the test set, while all the classes from the other domains represented the training dataset. This prevented \Sys from using classes within the same domain as test set to find the optimal subset. We report the average across $10$ different runs for each experiment within Table \ref{tab:add_results_table}. A similar trend to Table \ref{tab:results_table} appears within Table \ref{tab:add_results_table} where \Sys outperforms the baselines. Interestingly, \Sys maintained its dominance in performance across all experiments, indicating that the effects of the finite sample effect as discussed within Section \ref{sec:res:quant} may be limited in larger, more complex datasets. 

\input{figures/figure6}

\section{Additional figures}

\subsection{Pie-Charts}
\label{sec:pie_charts}
In the experimental section \ref{sec:res}, we provided a visualization of some of the class selections provided by our method for different datasets. In figure \ref{fig:pie:openset2} and figure \ref{fig:pie:disjoint:pacs}, we provide additional visualizations for the other datasets we conducted our experiments on. Specifically, for figure \ref{fig:pie:openset2} we provide a visualization of the class selection for Cifar-10 in an open set domain adaptation setting. \ref{fig:pie:openset2} provides the classes selected by WaSS on the PACS dataset within the disjoint domain adaptation setting.

\input{figures/figure2}
\subsection{UMAP}
\label{para:vis_res}

\autoref{fig:vis} show UMAP ~\citep{mcinnes2018umap-software} visualizations of specific adaptions of the Fashion-MNIST and CIFAR-10 datasets. Particularly, they illustrate the distribution of the training and target domains before and after we utilized WaSS for subset selection. We downsample the source domain to contain the same number of samples as the target domain. In the `before' example, the source domain is downsampled following the existing class distribution. In the `after' examples, we instead sample according to the class weighting distribution learned by WaSS.
\input{figures/figure10}

\input{figures/figure11}
The distance between the source (red dots) and target (black dots) domains is relatively smaller after plots enforcing that our method selects a mixture that brings the source domain closest to the target according to Wasserstein distance. For example, WaSS removes the largely unuseful classes on the left side of Fashion-MNIST from `before'. In CIFAR-10, the distribution of source classes is restricted to better conform to the shape of the target domain. 
\subsection{Statistical Tests}

\label{sec:stats}

We conduct paired T-tests to determine the statistical significance of the results shown in Figure \ref{fig:boxplot}. The following figures \ref{fig:fmnist_stats} \ref{fig:cifar10_stats} show the test statistic and p-value results for each pair of methods on Fashion-MNIST and CIFAR-10.

% \section{Classifier Hyperparameters}

% We use a ?? ResNet model to embed images into the latent space. Then, we learn a feed-forward network with ?? linearities and ??? hidden units and ?? learning rate and ?? optimizer. 

% We use the same hyperparameters across all transfer learning methodologies considered in our paper to ensure fairness of comparison. 

\input{jointMarginalProof}

\input{generaliztionBound}

%% file: tables/table4.tex
\begin{table}
  \caption{Accuracy of the downstream classifier trained on source distribution weighted by classes selected according to each baselines method in total class disjointness on Office-31 (O-31), Office-Home (OH), VizDa (VD). Best performing results are bolded.}
\label{tab:add_results_table}
\small
\resizebox{\columnwidth}{!}{
  \centering
  \begin{tabular}{llllllll}
%    \toprule
%    \cmidrule(r){1-2}
    Dataset     & Test Class  & All  & PADA & RND & MN & OSS &WASS  \\
    \midrule
O-31 & $[0,1,2]$ &  70.65 & 67.23 & 66.67 & 69.45 & 61.57 &\textbf{73.81} \\
O-31 & $[31,32,33]$ &  48.27 & 45.56 & 43.89 & 41.65 & 42.87 &\textbf{52.88} \\
O-31 & $[62,63,64]$ &  52.33 & 53.14 & 50.31 & 48.14 & 51.24 &\textbf{54.44} \\
\hline
OH   & $[0,1,2]$ &  53.14 &59.87 &58.67 &59.41& 53.28 &\textbf{61.88} \\
OH   & $[65,66,67]$ & 55.21  &52.87 &50.45 &54.36& 53.09 &\textbf{57.73} \\
OH   & $[130,131,132]$ &  69.84 &68.45 &67.53& 63.76& 61.29 & \textbf{73.54} \\
\hline
VizDa  & $[0,1,2]$ &  78.12 &76.89 &77.23 &75.34& 75.57  &\textbf{81.75} \\
VizDa  & $[12,13,14]$ &  63.56 &60.87 &59.78 & 60.12 & 62.12 &\textbf{64.38} \\
VizDa  & $[24,25,26]$ &  75.61 & 77.21 & 76.78 & 75.47 & 76.91  &\textbf{78.47} \\

% \hline
% Average  &  & 63.95 & 58.78 & 54.35 & 56.78 & 55.03 & \textbf{66.84} \\
    \bottomrule
  \end{tabular}
 }

\end{table}

%% file: figures/figure7.tex
\begin{figure}[t!]
\centering
	\begin{subfigure}[]{0.5\textwidth}
		\includegraphics[width=\textwidth]{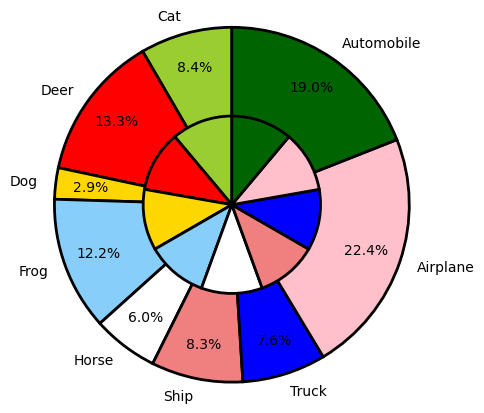}
		\caption{WaSS}
		\label{fig:pie:openset:cifar10:wass}
	\end{subfigure}
	\begin{subfigure}[]{0.49\textwidth}
		\includegraphics[width=\textwidth]{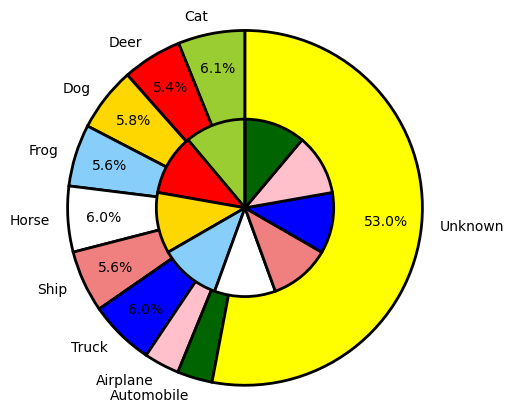}
		\caption{OSS}
		\label{fig:pie:openset:cifar10:oss}
	\end{subfigure}
    \caption{Class distributions for Cifar-10 of test set (inner circle) and class distribution of training set (outer circle) weighted by class selection methods: WaSS and OSS. The same color corresponds to the same class.}
    \label{fig:pie:openset2}
\end{figure}

%% file: figures/figure6.tex
\begin{figure}[t!]
\centering
\includegraphics[width=0.5\textwidth]{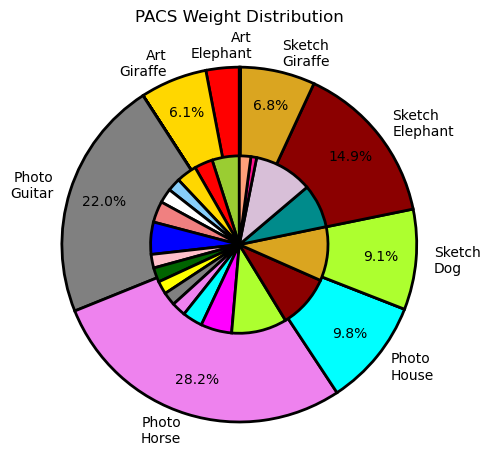}
\caption{Class distributions for PACS before (inner circle) and after (outer circle) applying our method to arbitrarily selected test classes. The same color corresponds to the same class.}
\label{fig:pie:disjoint:pacs}
\end{figure}

%% file: figures/figure2.tex
\begin{figure}[t!]
\centering
    \begin{subfigure}[]{1.0\columnwidth}
    	\caption*{Before}
    	\begin{subfigure}[]{0.49\textwidth}
    		\includegraphics[width=\textwidth]{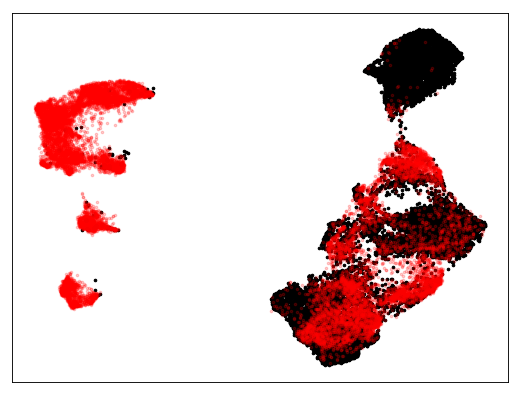}
    		%\caption{Fashion-MNIST}
    		\label{fig:vis:vis_before:mnist}
    	\end{subfigure}
    	\begin{subfigure}[]{0.49\textwidth}
    		\includegraphics[width=\textwidth]{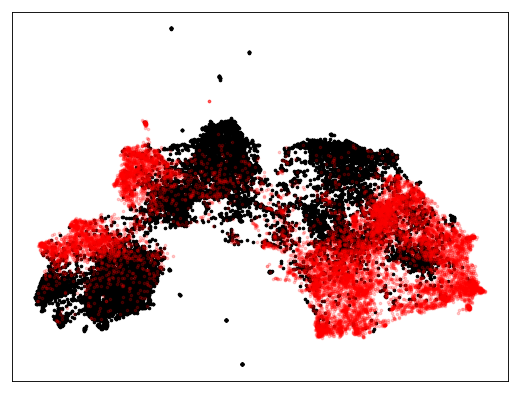}
    		%\caption{CIFAR 10}
    		\label{fig:vis:vis_before:cifar10}
    	\end{subfigure}
    \end{subfigure}
	\begin{subfigure}[]{1.0\columnwidth}
    	\caption*{After}
    	\begin{subfigure}[]{0.49\textwidth}
    		\includegraphics[width=\textwidth]{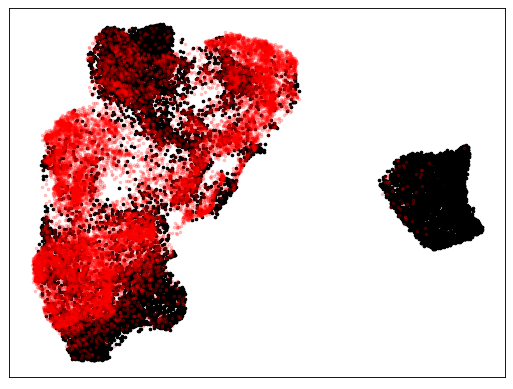}
    		\caption{Fashion-MNIST}
    		\label{fig:vis:vis_after:mnist}
    	\end{subfigure}
    	\begin{subfigure}[]{0.49\textwidth}
    		\includegraphics[width=\textwidth]{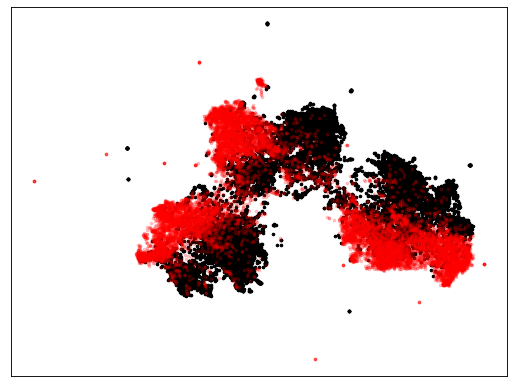}
    		\caption{CIFAR 10}
    		\label{fig:vis:vis_after:cifar10}
    	\end{subfigure}
	\end{subfigure}
	%\begin{subfigure}[]{0.32\textwidth}
	%	\includegraphics[width=\textwidth]{figures/pacs_before.png}
	%	\caption{PACS}
	%	\label{fig:vis_before:pacs}
	%\end{subfigure}
    \caption{A visualization showing the UMAP representations of the image features before (top row) and after (bottom row) application of WaSS. See \autoref{para:vis_res} for detailed explanation.}
    \label{fig:vis}
\end{figure}

%% file: figures/figure10.tex
\begin{figure}[t!]
\centering
	\begin{subfigure}[]{1.0\columnwidth}
		\includegraphics[width=\columnwidth]{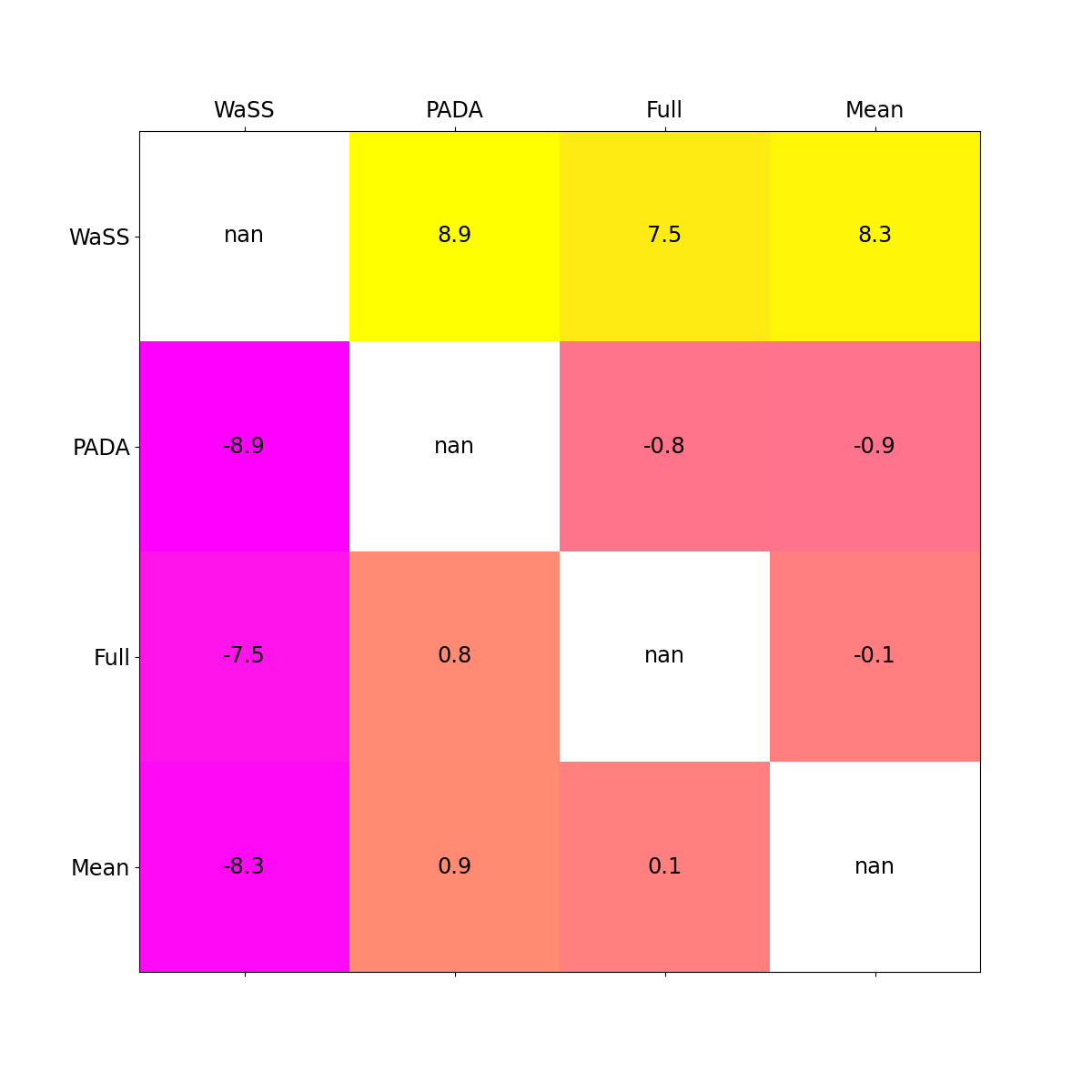}
		\caption{Test Statistic}
	\end{subfigure}
	\begin{subfigure}[]{1.0\columnwidth}
		\includegraphics[width=\columnwidth]{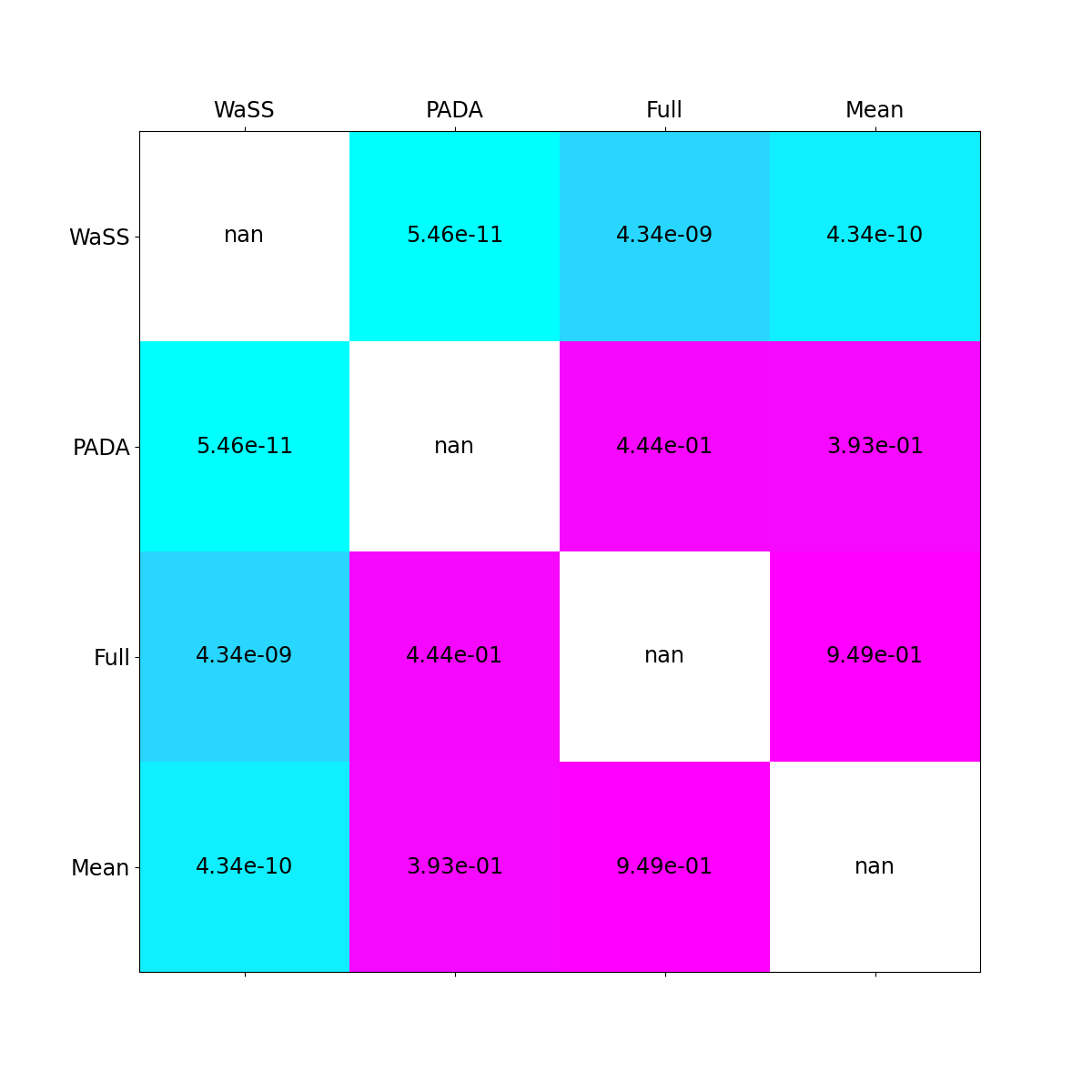}
		\caption{P-Value}
	\end{subfigure}
    \caption{Fashion-MNIST Statistical Tests}
    \label{fig:fmnist_stats}
\end{figure}

%% file: figures/figure11.tex
\begin{figure}[t!]
\centering
	\begin{subfigure}[]{1.0\columnwidth}
		\includegraphics[width=\columnwidth]{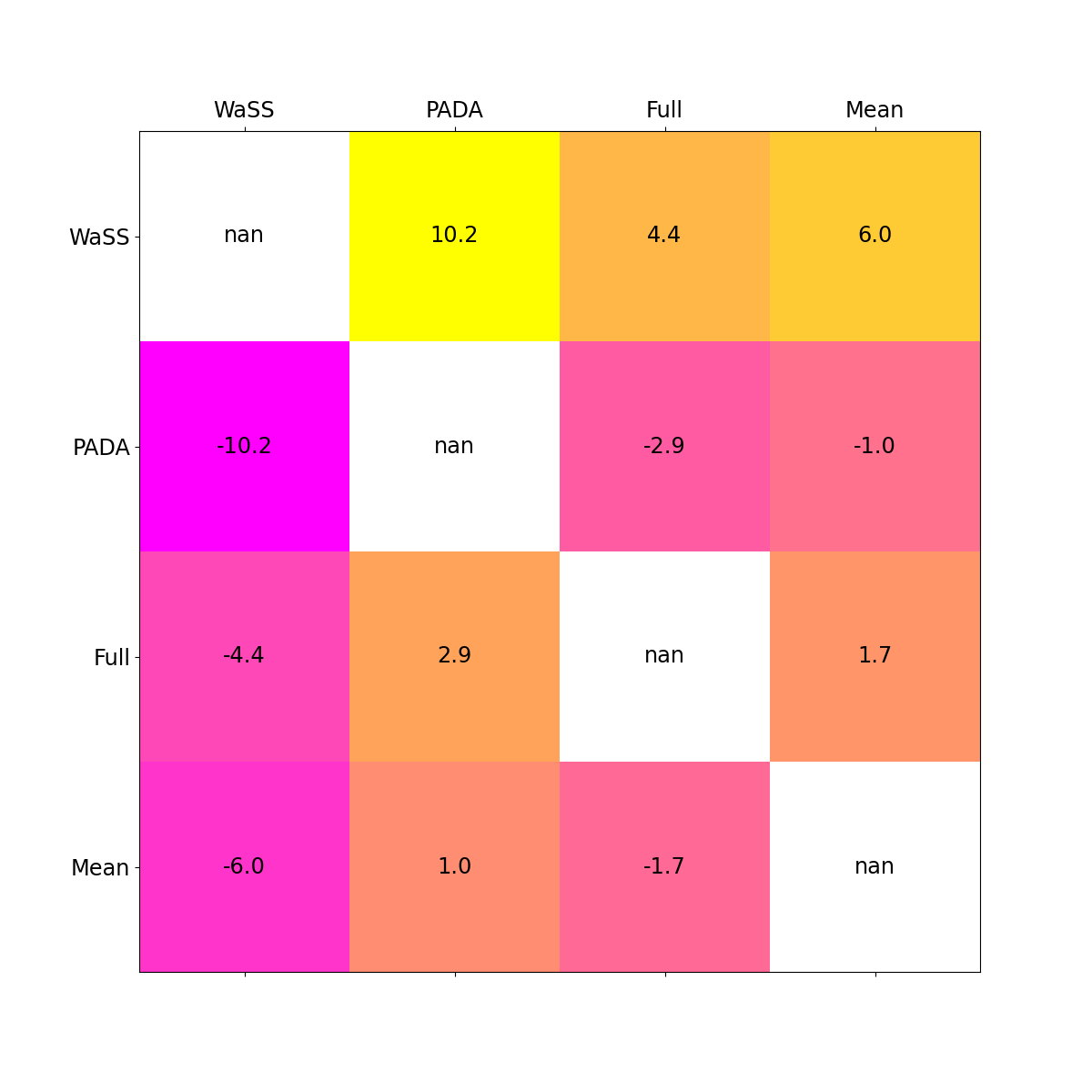}
		\caption{Test Statistic}
	\end{subfigure}
	\begin{subfigure}[]{1.0\columnwidth}
		\includegraphics[width=\columnwidth]{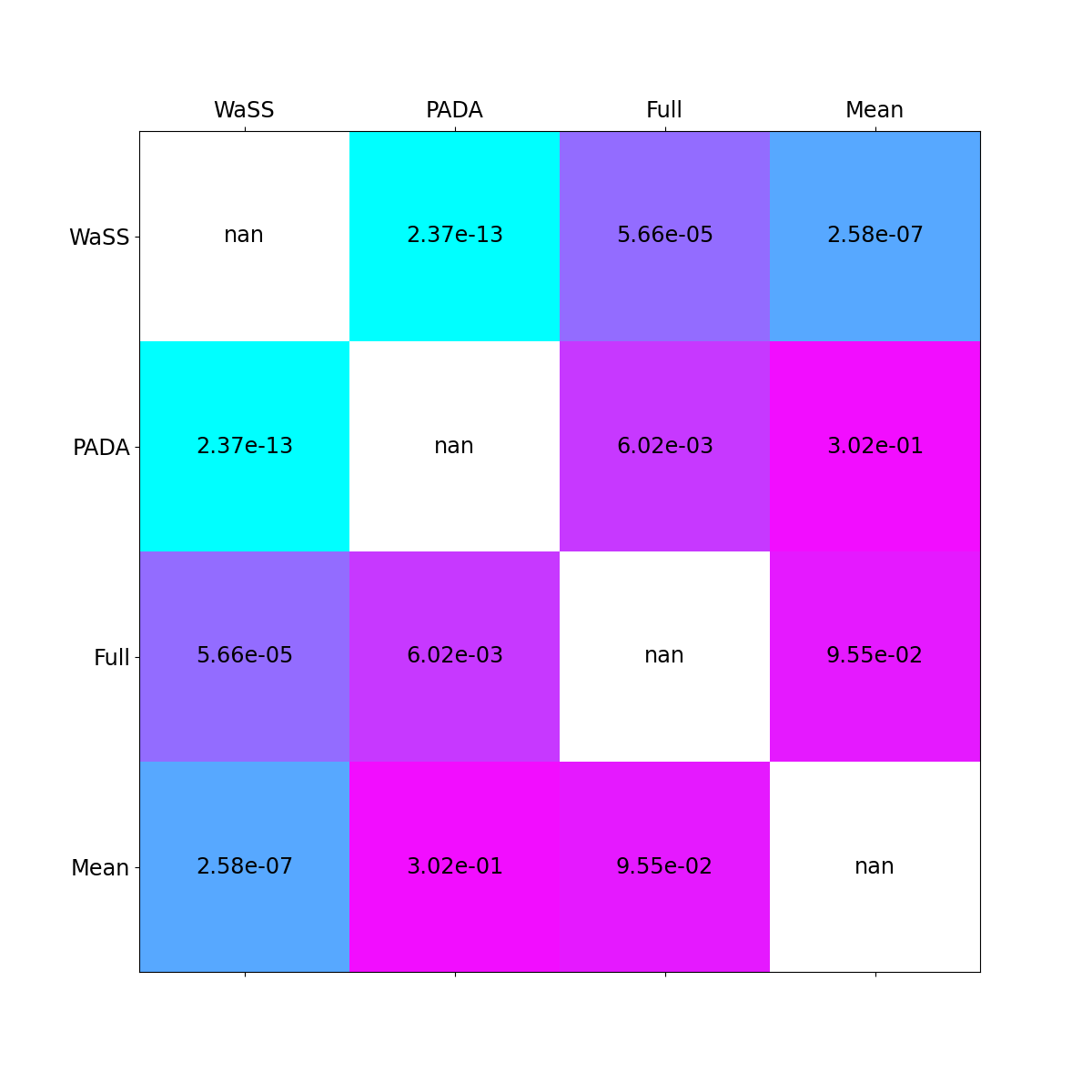}
		\caption{P-Value}
	\end{subfigure}
    \caption{CIFAR-10 Statistical Tests}
    \label{fig:cifar10_stats}
\end{figure}

%% file: jointMarginalProof.tex
\section{Proof of Lemma \ref{lemma:decomp}}
\textbf{Lemma \ref{lemma:decomp} (Error bound of Wasserstein distance between joint distributions)} 

\emph{For any two joint distributions $\gD_S$ and $\gD_T$ over $\gZ\times\gY$, we have}
    \begin{align*}
        W_1(\gD_S, \gD_T) &\leq W_1(\gD_S(Z), \gD_T(Z)) \\
        &+ \min\left\{\E_{\gD_S(Z)}[W_1(\gD_S(Y\mid Z), \gD_T(Y\mid Z)]\right.,\\ 
        &\hspace{3.2em}\left.\E_{\gD_T(Z)}[W_1(\gD_S(Y\mid Z), \gD_T(Y\mid Z)]\right\}.
    \end{align*}

\emph{Proof:} Define $\mu_M$ as the intermediate joint distribution over $Z \times Y$ such that $\mu_M = \mu_S(Z) \cdot \mu_T(Y | Z)$. \footnote{Note a result of this construction is that the intermediate input space is the equivalent between $\mu_M, \mu_S$ or $Z_M = Z_S$, not just that $\mu_M(Z) = \mu_S(Z).$} As Wasserstein distance is a metric:
\begin{align}
    W_1(\mu_S, \mu_T) \leq W_1(\mu_S, \mu_M) + W_1(\mu_M, \mu_T) 
\end{align}
This proof will upperbound $W_1(\mu_S, \mu_M)$ and $W_1(\mu_M, \mu_T)$, we start off by upperbounding $[W_1(\mu_S, \mu_M)]$.
Under definition (3) in \citep{courty2017joint} for joint distributions $\mathbb S \subseteq \R^d$ and $\mathbb C$ (the set of labels):
\begin{align}
\label{def:jointWass}
    W_1(\mu_S, \mu_M) = \inf _{\gamma \in \Gamma(\mu_S, \mu_M)} \int_{{\mathbb S \times \mathbb C}^2} &\mathcal{D}(z_m, y_m; z_s, y_s) \\
    &\mathrm{~d} \gamma(z_m, y_m; z_s, y_s) \nonumber 
\end{align}
\citep{courty2017joint} defined $\mathcal{D}(z_m, y_m; z_s, y_s) = \alpha d(z_m, z_s) + \mathcal{L}(y_m, y_s)$ where $d(z_m, z_s)$ and $\mathcal{L}(y_m, y_s)$ are distance functions between $z_m, z_s$ and  $y_m, y_s$ respectively\footnote{For our setting, we assume $d_Z:Z\times Z \rightarrow \R^d$ and $\mathcal{L}_Y:Y\times Y \rightarrow \R^d$.}.  
Since $Z_m, Z_S$ are equal $d(z_m, z_s) = 0$ and \ref{def:jointWass} can be rewritten as:
\begin{align}
\label{def:jointWass3}
    &W_1(\mu_S, \mu_M) = \nonumber\\
    &=\inf _{\gamma \in \Gamma(\mu_S, \mu_M)} \int_{({\mathbb S \times \mathbb C})^2} \mathcal{L}(y_m, y_s) \mathrm{~d} \gamma(z_m, y_m; z_s, y_s)  \\
\label{def:jointWass4}
    &= \inf _{\gamma \in \Gamma(\mu_S, \mu_M)} \int_{({\mathbb S \times \mathbb C})^2} \mathcal{L}(y_m, y_s)\mathrm{~d} \gamma(y_m|z_m; y_s|z_s)\\
    &\hspace{8em}  * \mathrm{~d}\gamma(z_m, z_s) \nonumber \\
\label{def:jointWass5}
    &= \inf _{\gamma \in \Gamma(\mu_S, \mu_T)} \int_{({\mathbb S \times \mathbb C})^2} 
    \mathrm{~d}\gamma(z_m, z_s) \\
    &\hspace{4.5em} \int_{({\mathbb S \times \mathbb C})^2} \mathcal{L}(y_m, y_s) \mathrm{~d} \gamma(y_m|z_m=z; y_s|z_s=z) \nonumber 
\end{align}
\ref{def:jointWass3} can be reduced to \ref{def:jointWass4} because of law of total probability (full proof in \ref{helper1}). Given distance in \ref{def:jointWass4} is only calculated with respect to $Y$ we can fix $Z$  allowing us to transform \ref{def:jointWass4} to \ref{def:jointWass5} given $\mu_M$ is defined as $\mu_S(Z) \cdot \mu_T(Y | Z)$. 

Given $(z,z')$ are fixed pairs, $\mathrm{~d}\gamma(z_m, z_s)$ can be brought out from the infimum:
\begin{align}
\label{def:jointWass6}
    &=  \int_{({\mathbb S \times \mathbb C})^2} \mathrm{~d}\gamma(z_m, z_s) \inf _{\gamma \in \Gamma(\mu_S(Y|Z_s=z), \mu_T(Y|Z_m=Z))}\int_{({\mathbb S \times \mathbb C})^2}\\
    &\mathcal{L}(y_m, y_s) \mathrm{~d} \gamma(y_m|z_m=z; y_s|z_s=z) \nonumber
\end{align}

 Based on our definition of  $\mu_M$, \ref{def:jointWass6} is nothing but the Wasserstein distance between $\mu_S, \mu_T$:
\begin{align}
\label{def:jointWass7}
    &=  \int_{({\mathbb S \times \mathbb C})^2} \mathrm{~d}\gamma(z_m, z_s)  W_1(\mu_S(Y|Z=z), \mu_T(Y|Z=z))\\
\label{def:jointWass8}
    &=  \mathop{\mathbb{E}}_{\mu_{S}(Z)} [W_1(\mu_S(Y|Z=z), \mu_T(Y|Z=z))]
\end{align}
where \ref{def:jointWass8} is nothing but the definition of expectation. Now we consider upper bounding $[W_1(\mu_M, \mu_T)]$. 

Based on \ref{def:jointWass}, we can define $[W_1(\mu_M, \mu_T)]$ as:
\begin{align}
\label{def:jointWass9}
    W_1(\mu_M, \mu_T) = \inf _{\gamma \in \Gamma(\mu_M, \mu_T)} &\int_{({\mathbb S \times \mathbb C})^2} d(z_m, z_t) 
    \\&+ \mathcal{L}(y_m, y_t) \mathrm{~d} \gamma(z_m, y_m; z_t, y_t)\nonumber
\end{align}
\begin{align}
\label{def:jointWass10}
     =\inf _{\gamma \in \Gamma(\mu_M, \mu_T)} &\int_{({\mathbb S \times \mathbb C})^2} d(z_m, z_t) \gamma(z_m, y_m; z_t, y_t) \\
     &+ \int_{({\mathbb S \times \mathbb C})^2}\mathcal{L}(y_m, y_t) \mathrm{~d} \gamma(z_m, y_m; z_t, y_t) \nonumber
\end{align}
\ref{def:jointWass10} is a simple distribution of the $\gamma$ term. 
Given both distance functions are isolated, we rewrite each part of the equation similar to \ref{def:jointWass4}. 
\begin{align}
\label{def:jointWass11}
    &=\inf _{\gamma \in \Gamma(\mu_M, \mu_T)} \int_{({\mathbb S \times \mathbb C})^2} d(z_m, z_t) \gamma(z_m, z_t) \gamma(y_m|z_m; y_t|z_t) \\
    &\hspace{5em}+   \int_{({\mathbb S \times \mathbb C})^2} \mathcal{L}(y_m, y_t) \mathrm{~d} \gamma(z_m,z_t) \gamma(y_m|z_m; y_t|z_t) \nonumber\\
\label{def:jointWass18}
    &=\inf _{\gamma \in \Gamma(\mu_M, \mu_T)} \int_{({\mathbb S \times \mathbb C})^2} \gamma(y_m|Z=z; y_t|Z=z) \\
    &\hspace{5em}\int_{({\mathbb S \times \mathbb C})^2} d(z_m, z_t) \gamma(z_m, z_t)\nonumber \\ 
    &\hspace{5em} +\int_{({\mathbb S \times \mathbb C})^2}  \mathcal{L}(y_m, y_t) \mathrm{~d} \gamma(z_m,z_t) \gamma(y_m|z_m; y_t|z_t) \nonumber\\
\label{def:jointWass12}
    &=\inf _{\gamma \in \Gamma(\mu_M, \mu_T)} \int_{({\mathbb S \times \mathbb C})^2} d(z_s, z_t) \gamma(z_s, z_t) \\ 
    &\hspace{5em} + \int_{({\mathbb S \times \mathbb C})^2} \gamma(z_m,z_t) \nonumber \\
    &\hspace{5em}\int_{({\mathbb S \times \mathbb C})^2} \mathcal{L}(y_m, y_t)\mathrm{~d} \gamma(y_m|z_m; y_t|z_t) \nonumber
\end{align}
where \ref{def:jointWass12} uses the definition of $\mu_M$ to cancel out $\gamma(y_m|Z=z; y_t|Z=z)$ and change $z_m$ to $z_s$ in the first part of the equation.  For the second part of the equation, under the definition of $\mu_M$, we know that the conditional distribution of $\mu_M$ and $\mu_T$ are the same. This means that the output space between the two distributions is also the same allowing $\mathcal{L}(y_m, y_t)$ can be rewritten as $\mathcal{L}(y_t, y_t) = 0$ which results:  
\begin{align}
\label{def:jointWass13}
    =\inf _{\gamma \in \Gamma(\mu_S(Z), \mu_T(Z))} \int_{({\mathbb S \times \mathbb C})^2} d(z_s, z_t) \gamma(z_s, z_t) 
\end{align}
\begin{align}
\label{def:jointWass14}
    W_1(\mu_M, \mu_T) = W_1(\mu_S(Z), \mu_T(Z))
\end{align}
 where \ref{def:jointWass14} incorporates the definition of Wasserstein.

Combining \ref{def:jointWass8} and \ref{def:jointWass14}, we get the following:

\begin{align}
\label{def:jointWass15}
    W_1(\mu_S, \mu_T) \leq &W_1(\mu_S(Z) , \mu_T(Z)) \\
    &+  \mathop{\mathbb{E}}_{\mu_{S}(Z)}[W_1(\mu_S(Y|Z=z), \mu_T(Y|Z=z))] \nonumber
\end{align}

If $\mu_M = \mu_T(Z) * \mu_S(Y|Z)$, under the same logic, \ref{def:jointWass15} could be rewritten as:

\begin{align}
\label{def:jointWass16}
    W_1(\mu_S, \mu_T) \leq &W_1(\mu_S(Z) , \mu_T(Z)) \\
    &+  \mathop{\mathbb{E}}_{\mu_{T}(Z)}[W_1(\mu_S(Y|Z=z), \mu_T(Y|Z=z))] \nonumber
\end{align}

This results in:
\begin{align}
\label{def:jointWass17}
    W_1(\mu_S, \mu_T) \leq & W_1(\mu_S(Z) , \mu_T(Z)) + \\ &\min\{\mathop{\mathbb{E}}_{\mu_{T}(Z)}[W_1(\mu_S(Y|Z=z), \mu_T(Y|Z=z))]+ \nonumber\\ &\mathop{\mathbb{E}}_{\mu_{T}(Z)}[W_1(\mu_S(Y|Z=z), \mu_T(Y|Z=z))]\} \nonumber
\end{align}

\subsubsection{Helper Proof 1:}
\label{helper1}
Prove: $\gamma(z_m, y_m; z_s, y_s) = \gamma(y_m|z_m; y_s|z_s) \gamma(z_m, z_s)$
\begin{align}
    \label{helpProof:1}
    \gamma(z_m, y_m; z_s, y_s) &= \gamma(z_m, z_s, y_m, y_s) \\
     \label{helpProof:2}
    &= \gamma(z_m, z_s) * \gamma(y_m, y_s| z_m, z_s) \\
      \label{helpProof:3}
     &= \gamma(z_m, z_s) * \gamma(y_m|z_m, y_s|z_s) 
\end{align}
Given every term in $\gamma$ is a random variable, we are able to shift around terms within $\gamma$ resulting in \ref{helpProof:1}. Next, the law of total probability, allow us to transform \ref{helpProof:1} to \ref{helpProof:2}. Finally, we know that for fixed pair of points, $\gamma(y_m, y_s| z_m, z_s)$ is just a coupling between $\gamma(y_m|z_m, y_s|z_s)$ which allows us to resolve \ref{helpProof:2} to \ref{helpProof:3}.

%% file: generaliztionBound.tex
\section{Proof of Lemma \ref{lemma:transfer}}

\textbf{Lemma \ref{lemma:transfer} (Error bound on pre-trained and fine-tuned model)} 

Given two classifiers $h,h'$ that only differ by their final layer, we can bound their error as follows:
\begin{align*}
   \epsilon_T(h') \leq \epsilon_S(h) + \max\{\rho, 1\}\cdot W_1(\gD_S, \gD_T) + \alpha \beta \sigma_{\max} (V_{S}-V_{T})
\end{align*}

\begin{proposition}
\label{prop:lipConst}
   There exists a Lipschitz Constant $\alpha$ for the Softmax function.
\end{proposition}

\begin{proposition}
\label{prop:l2norm}
  The $L_2$ norm of a vector is bounded by its largest singular value. 
\end{proposition}

\emph{Proof:} In an input space $X \subset \mathbb{R}^d$, define two classifiers, $h = OW(x)$ and $h = ZW(x)$, that differ in their final classification layer represented as matrices $O, Z$. Let $W: X \to F$ be the shared featurerizer where every resulting feature vector is bounded such that $||W(x)||_2 \leq \beta$ $\forall x \in X$. From Proposition \ref{prop:diff}, for two distributions $\gD_S, \gD_T$ we know 
\begin{align}
\epsilon_T(h^o) &\leq \epsilon_S(h^o) + \max\{\rho, 1\}\cdot W_1(\gD_S, \gD_T) \\ 
\epsilon_T(h^o) &\leq \epsilon_S(h^o) + \max\{\rho, 1\}\cdot W_1(\gD_S, \gD_T) \\
&\hspace{3.8em} + (\epsilon_S(h^*) - \epsilon_S(h^*))\nonumber\\
\epsilon_T(h^o) &\leq \epsilon_S(h^*) + \max\{\rho, 1\}\cdot W_1(\gD_S, \gD_T) \\
&\hspace{3.8em} + (\epsilon_S(h^o) -\epsilon_S(h^*)) 
\end{align}

Consider the final term $\epsilon_S(h^o) -\epsilon_S(h^*)$ on the right hand side. With labeling function $Y : X \to \mathbb{R}^k$:
\begin{align}
\epsilon_S(h^o) -\epsilon_S(h^*) &= \E X_S ||Y(x) - \sigma(OW(x))||_1  \\
&\hspace{0.5em}-\E X_S||Y(x) - \sigma(ZW(x))||_1 \nonumber\\
&= \E_{X_S} \left[ ||Y(x) - \sigma(OW(x))||_1\right. \\
&\left.\hspace{3.0em}-||Y(x) - \sigma(ZW(x))||_1 \right] \nonumber\\
&\leq \E_{X_S} \left[ ||Y(x) - \sigma(OW(x))\right.\\
&\hspace{3.0em}\left.- Y(x) + \sigma(ZW(x))||_1 \right] \nonumber \\
&= \E X_S ||\sigma(ZW(x)) - \sigma(OW(x))||_1 
\end{align}

Given Proposition \ref{prop:lipConst}, we know the softmax function $\sigma$ is bounded by a Lipshitz constant allowing for:
\begin{align}
\label{def:softmax:reduce}
    \E_{X_S} \left[ ||\sigma(ZW(x)) - \sigma(OW(x))||_1 \right] &\leq \\
    \alpha ||Z&W(x)-OW(x)||_2 \nonumber
\end{align}

Furthermore, given the features space is bounded, \ref{def:softmax:reduce} can be further reduced to:
\begin{align}
    \alpha ||(Z-O)W(x)||_2 \leq \alpha ||Z-O||_2 \beta
\end{align}

Using Proposition \ref{prop:l2norm}, $Z-O$ can be bounded by its largest singular value $\sigma_{\max}$,
\begin{align}
    \alpha \beta ||(Z-O)||_2 \leq \alpha \beta \sigma_{\max} (Z-O)
\end{align}

This results in:
\begin{align}
    \label{def:wass:transfer}
    \epsilon_T(h^o) \leq \epsilon_S(h^*) + \max\{\rho, 1\}\cdot W_1(\gD_S, \gD_T)\\
    \hspace{4.5em} +  \alpha \beta \sigma_{\max} (Z-O)\nonumber
\end{align}

If no finetuning is done ($h^o = h^* \iff Z = O$), \ref{def:wass:transfer} collapses to Proposition \ref{prop:diff}.